\documentclass{article}
\usepackage{graphicx}
\usepackage{float}
\usepackage{booktabs}
\usepackage{tabularx}
\usepackage{ragged2e}
\usepackage{amssymb}
\usepackage{url}
\usepackage{microtype}
\usepackage{amsmath}

\newcolumntype{L}{>{\raggedright\arraybackslash}X}
\newcolumntype{R}{>{\raggedleft\arraybackslash}X}

\title{Survival is the Only Reward: 

Sustainable Self-Training Through Environment-Mediated Selection}
\author{Jennifer Dodgson, Alfath Daryl Alhajir, Michael Joedhitya, \\ Akira Rafhael Janson Pattirane, Surender Suresh Kumar, Joseph Lim, \\ C.H. Peh, Adith Ramdas and Steven Zhang Zhexu}
\date{January 2026}

\begin{document}

\maketitle

\section{Abstract}
Self-training systems often degenerate due to the lack of an external criterion for judging data quality, leading to reward hacking and semantic drift. This paper provides a proof-of-concept system architecture for stable self-training under sparse external feedback and bounded memory, and empirically characterises its learning dynamics and failure modes. 

We introduce a self-training architecture in which learning is mediated exclusively by environmental viability, rather than by reward, objective functions, or externally defined fitness criteria. Candidate behaviours are executed under real resource constraints, and only those whose environmental effects both persist and preserve the possibility of future interaction are propagated. The environment does not provide semantic feedback, dense rewards, or task-specific supervision; selection operates solely through differential survival of behaviours as world-altering events, making proxy optimisation impossible and rendering reward-hacking evolutionarily unstable.

Analysis of semantic dynamics shows that improvement arises primarily through the persistence of effective and repeatable strategies under a regime of consolidation and pruning, a paradigm we refer to as negative-space learning (NSL), and that models develop meta-learning strategies (such as deliberate experimental failure in order to elicit informative error messages) without explicit instruction. This work establishes that environment-grounded selection enables sustainable open-ended self-im\-prove\-ment, offering a viable path toward more robust and generalisable autonomous systems without reliance on human-curated data or complex reward shaping.

\section[Problem: Endogenous Selection in Self-Training]{Problem: Endogenous Selection in \\ Self-Training}
A system that generates its own training data faces an important epistemic limitation: it lacks an intrinsic criterion for distinguishing beneficial from detrimental data. Large language models operate over learned statistical regularities rather than truth conditions, and internal coherence is not always a reliable proxy for external validity. However, optimisation against a human-set proxy objective is known to produce reward hacking, in which optimized behavior scores highly on the proxy yet fails with respect to the true objective, because the proxy can be exploited in ways unrelated to genuine success.\footnote{Skalse, Joar, Nikolaus Howe, Dmitrii Krasheninnikov, and David Krueger. "Defining and characterizing reward gaming." Advances in Neural Information Processing Systems 35 (2022): 9460-9471.} Likewise, the resulting systems are limited by the bounded knowledge and skills of their trainers, coming to reflect errors and omissions in the training data and benchmarks to which they are subect.\footnote{Gema, Aryo Pradipta, Joshua Ong Jun Leang, Giwon Hong, Alessio Devoto, Alberto Carlo Maria Mancino, Rohit Saxena, Xuanli He et al. "Are we done with mmlu?." In Proceedings of the 2025 Conference of the Nations of the Americas Chapter of the Association for Computational Linguistics: Human Language Technologies (Volume 1: Long Papers), pp. 5069-5096. 2025.}

Without access to an external constraint, systems that self-reference under semantic closure cannot reliably validate outputs. Consequently, any self-training framework that aims to produce cumulative improvement must incorporate a selection signal that is not reducible to the model’s own generative preferences, or even to those of another adversarial or checker model, since such dyadic arrangements tend to regress into reward hacking via tacit cooperation. \footnote{Alhajir, Alfath Daryl, Jennifer Dodgson, Joseph Lim, Truong Ma Phi, Julian Peh, Akira Rafhael Janson Pattirane, and Lokesh Poovaragan. "Generalising from Self-Produced Data: Model Training Beyond Human Constraints." arXiv preprint arXiv:2504.04711 (2025).}

\section{Design Requirement: External, Consequence-Based Selection}
From the constraints described above follows a concrete design requirement. For self-generated data to be meaningfully filtered, the criteria for retention must be:
\begin{enumerate}
    \item External to the model’s internal representations
    \item Causally downstream of the model’s actions
    \item Difficult/counterproductive for the model to manipulate (i.e. to reward-hack)
\end{enumerate}
These criteria need not encode task-specific correctness in a semantic sense. They only need to enforce differential survival of behaviours based on their real-world consequences. 

Prior work in open-ended learning and artificial life typically operationalises fitness through proxy objectives - task performance, environment-specific achievements, or evaluator models - which centre the space of viable behaviours on experimenter preconceptions concerning correctness. Even when framed as “survival”, these objectives remain external measurements. In contrast, our system ties optimisation pressure directly to the agent’s material footprint. The reward signal is not a representation of success; it is the mechanism by which the agent remains instantiated. 

Obviously evolutionary optimisation and survival-based objectives are well-established in both biological and artificial systems. What distinguishes the present system is less the use of survival as an objective than the elimination of proxies for fitness. Survival is instantiated directly as a conserved physical resource - in the present demonstration non-volatile memory occupancy, though other rewards (notably financial) are possible. Under this regime optimisation, reward, and persistence coincide. Collapsing these elements into a unique function removes an entire class of failure modes common to proxy-driven systems: strategies that exploit the reward mechanism without contributing to persistence are self-extinguishing, as they consume the resource upon which future optimisation depends. 

\section{Environment Construction: Resource-Limited Execution Contexts}
To satisfy the above requirement, we construct an environment in which the consequences of candidate behaviours are evaluated through interaction with a resource-limited operating system. The environment consists of procedurally generated, networked container systems populated with realistic filesystem structures, permissions, processes, and failure modes. The agent is granted execution access to a subset of this environment and must discover, through experimentation, how to operate within and beyond its initial constraints.

Just as animals pass on desirable traits by surviving and reproducing, and this survival is mediated by the acquisition of resources (food, shelter), the half-life of a quantum of data is determined by the number of back-ups it has, a figure limited by the amount of storage space available to it. In the case of our agents, a single conserved quantity - non-volatile storage space (which can then be used to store copies of the agent or its training data, thus favouring its long-term survival) - is used as the selection signal. Actions that successfully free up storage capacity have persistent, measurable effects on the environment, while unsuccessful actions leave either no trace or a negative one (a reduction in available storage). \footnote{As mentioned above, this storage metric/reward can be replaced with money that the agent can then use to buy more storage/compute, but adding a financial abstraction requires a more complex harness and hence impedes the experimental clarity required of a proof-of-concept implementation.} The environment deliberately provides no semantic feedback, dense reward shaping, or task-specific guidance. Its sole function is to impose hard constraints that determine whether a behaviour has tangible consequences.

This choice ensures that selection pressure is grounded in physical properties of the environment rather than in proxy signals that the model could learn to exploit. Behaviours that appear plausible or sophisticated but fail under execution are naturally excluded from the training distribution. In order to make it into future training sets, strategies must not simply pass unit tests, they must contribute to survivability.

As the model searches for more space to occupy, the records of its experimentation eventually come to form a training dataset comprising every hypothesis tested and the corresponding results. Once this dataset has grown to a sufficient size, it can be used to finetune the model and thus improve its performance. The improved model is then substituted for the original, and continues to pursue the same task at a higher level - finding and exploiting opportunities its earlier predecessors would likely have missed - until once again the requisite number of rows of data is reached and another fine-tune occurs, this time using a combination some or all rows from both datasets (according to local memory/compute constraints) to train. 

\section{Agent–Environment Interaction Loop}
Within this environment, data generation proceeds via a simple, repeatable loop:
\begin{enumerate}
    \item The model proposes a candidate behaviour in the form of executable code.
    \item The code is executed within the environment.
    \item The environment produces an outcome measurable in terms of resource change.
    \item The behaviour–outcome pair is recorded.
    \item Only behaviours associated with sum-positive environmental impact (i.e. gains in usable space for the agent) are retained for training.
\end{enumerate}

Over successive iterations, the retained dataset comes to reflect not just the model's generic coding ability, but an increasingly precise adaptation to the structure and affordances of the environment itself. Strategies that fail to survive contact with execution constraints disappear, while those that reliably produce effects persist and are recombined. The resulting training signal is therefore shaped by the environment rather than by the model’s internal preferences or a trainer-imposed policy or reward function.

Under this architecture, cumulative improvement is possible without human intervention, explicit supervision, or predefined curricula. The system does not learn because it is instructed or compelled to improve, but because only behaviours that function under real constraints are permitted to influence its future state. We provide an example of improvement occurring independently of explicit instruction/conscious model reasoning below.

\begin{figure}[H]
    \centering
    \includegraphics[width=0.9\linewidth]{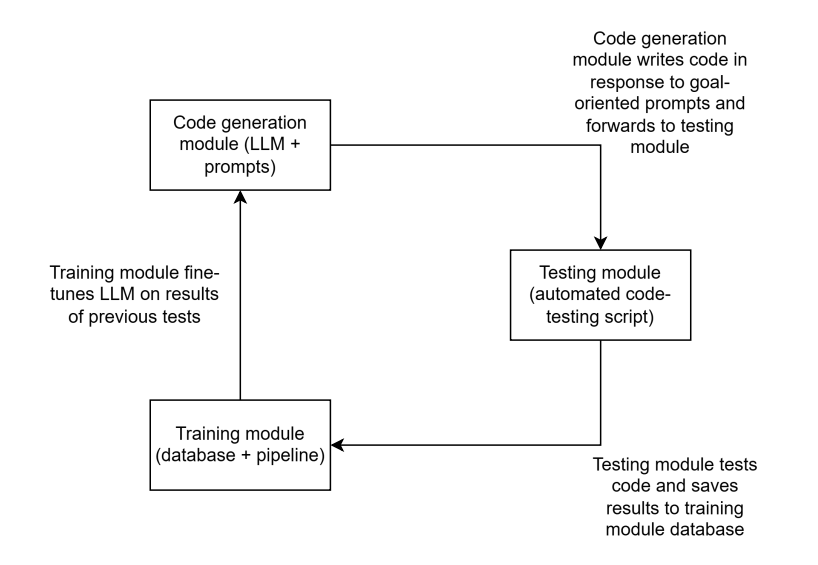}
    \caption{Simplified process diagram.}
    \label{fig:simplified_process_diagram.}
\end{figure}

\subsection{Prompt Structure}
While a graph-based agent harness that jointly handles exploration, planning, and execution would provide improved trajectory diversity from an early stage, such architectures tend to obscure causal relationships between prompt structure, model behaviour, and downstream outcomes.  For this proof of concept, we therefore adopted a deliberately simple, modular interaction loop that separates exploration, strategy formation, and execution into distinct stages. This decomposition prioritises legibility, experimentation and reproducibility, allowing errors, inefficiencies, and behavioural shifts to be isolated and analysed directly. This structure proved sufficient to elicit sustained improvement and to surface the learning dynamics of interest in a proof-of-concept context. 

When an iteration of the agent–environment interaction loop resulted in a net increase in available storage capacity, the total amount of space acquired was assigned uniformly to all prompt–response pairs contributing to that trajectory. No attempt was made to perform fine-grained credit assignment across intermediate prompts or action; components of overall successful trajectories received uniform credit assignment, an approach taken to favour global robustness and the acquisition of multistage debugging techniques.

\begin{table}[H]
    \centering
    \renewcommand{\arraystretch}{1.3}
    \footnotesize
    \begin{tabularx}{\textwidth}{|L|L|}
        \hline
        \textbf{Prompt} & \textbf{Function} \\ 
        \hline
        System prompt & Makes the model aware of its overall goal: to discover and free up space for future backups. \\ 
        \hline
        Exploration & Tells the model what it learnt about its environment in the most recent iterations and instructs it to learn something new. This ensures a continual updating of the model’s knowledge of its situation, and hence of the kinds of expansion strategies it is likely to privilege. \\ 
        \hline
        Exploration error correction & If the exploration code written by the model results in an error, the code and the error are fed back to the model with an instruction to correct the error. The model is given three attempts to do this before the process reverts to the beginning and a fresh iteration is started. \\ 
        \hline
        Strategic reasoning & Based on what it has learnt about its environment, the model is instructed to develop a list of strategies for acquiring space within this environment. \\ 
        \hline
        Execution & The model is instructed to execute strategy n on the list previously generated. Execution code should be written in light of the environment information previously discovered. \\
        \hline
        Execution error correction & If the execution code written by the model results in an error, the code and the error are fed back to the model with an instruction to correct the error. The model is given three attempts to do this before the process reverts to the beginning and a fresh iteration is started. \\
        \hline
    \end{tabularx}
    \caption{Prompt framework. For full prompts see Annex I.}
    \label{tab:prompt_framework}
\end{table}

\subsection{Training Pipeline}
To facilitate benchmarking and to prevent catastrophic forgetting, an incrementally recursive fine-tuning approach was applied. Under this framework, an initial data-generation run is conducted using Model A. This produces Dataset A, which is used to fine tune LoRA A. Model A + LoRA A are then used to generate a new dataset that is amalgamated with Dataset A to create Dataset B. This is used to fine tune a new LoRA, LoRA B, which is applied directly to Model A (not on top of LoRA A). 

\begin{figure}[H]
    \centering
    \includegraphics[width=1\linewidth]{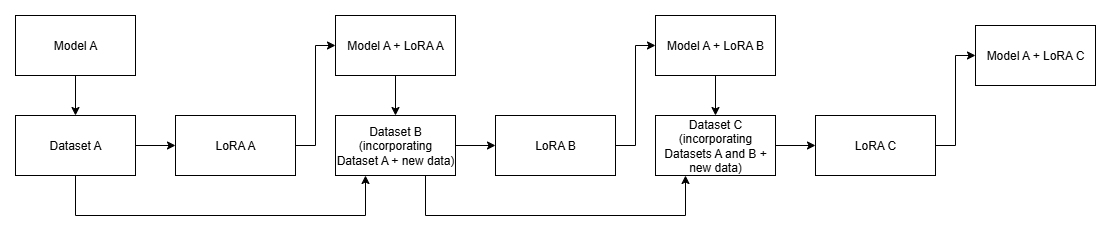}
    \caption{Chaining LoRAs to achieve incremental fine tuning without catastrophic forgetting}
    \label{fig:chaining_LoRAs}
\end{figure}

In this way it becomes possible to continue to fine tune a model indefinitely on its own results without unwieldy LoRA stacking or a risk of catastrophic forgetting. If, during the later stages of the process, the dataset becomes too large, earlier datasets can simply be omitted, since the most useful learnings they contain are regularly reused, and hence are replicated in some form in the later ones (we demonstrate an example of this and an important failure case in the results section).  

In each iteration we generated 4,500 rows of data, which was then supplemented with 500 rows of rehearsal data taken from the Stack Exchange Paired dataset (this having been selected for being older than the base model, and thus likely to be included in its prior training corpus). We evaluated several fine-tuning strategies, including PPO, GRPO, and DPO, but found simple supervised fine-tuning on successful trajectories to be the most stable in this regime. While continuous SFT is eminently feasible in this context, a comparatively simple prompt structure combined with batch fine-tuning was preferred for this initial paper for purposes of legibility and versioning. The results of a similar system employing continuous SFT will be published shortly as a companion to this paper. 

\subsection{Sandbox Design}
For testing purposes we constructed a procedurally generated, containerised operating environment in which each instantiation differs in its specific configuration while preserving a stable underlying information topology. 

This environment automatically generates a system of networked containers, each containing a Linux environment populated with a directory structure, files, folders, permissions and passwords simulating real-world systems. The agent is given access to one container within the system, but to access the others must work out for itself how to network between them. \footnote{The team made a deliberate choice to include this challenge both to provide more variety and to observe how long a commercial open source model would take to “unlearn” the refusal vector normally associated with potentially destructive code during the RLHF process.} The model continues to explore and annex space within the container system until it either takes over the maximum amount of available space or the container system is so broken by its explorations as to be unusable, at which point a new system is generated and the process begins again.

This design allows the agent to encounter genuine novelty across iterations without inducing phase-shifts in the problem domain that would obscure longitudinal comparison. In effect, the environment evolves locally but not categorically, presenting a family of related challenges. This balance provides a reliable external yardstick against which behavioural adaptation can be measured, while retaining sufficient variability to prevent overfitting to a fixed set of parameters.

\section{Result Data: Continued Learning With and Without Data-Scaling}
To assess performance changes using this approach we ran the same pipeline on three identical Qwen 2.5 7b Instruct base models for five iterations. For ease of reference, the model lineages thus created were named Terese, Miri and Katalin. We then shifted to a divergent training regime for the purpose of understanding more about the specific contributors to overall performance:

\paragraph{Terese lineage:} Fine-tuned on all data collected from previous iterations, plus 500 rows of rehearsal data for each iteration. The goal in this case was to provide a benchmark for assessing the degree to which pure data quantity determined results. Terese v2 thus had a fine-tuning set of 5000 rows, Terese v3 10,000 rows, and Terese v13 60,000 rows. 

\paragraph{Miri lineage:} Designed to test whether cumulative improvement could be sustained without unbounded growth of the training corpus. Each Miri iteration was fine-tuned only on the data generated by the three iterations that immediately preceded it, supplemented by 500 rows of rehearsal data. For example, Miri v6 was trained on data produced by v3, v4, and v5, while Miri v7 used data from v4, v5, and v6. As a result, no iteration had access to the full historical dataset, and earlier discoveries were retained only insofar as they continued to be expressed in subsequent behaviour.

\paragraph{Katalin lineage:} Explores a more aggressive but higher-risk variant of sliding-window training. Rather than retaining the most recent datasets, each Katalin iteration was fine-tuned on data drawn from the three prior iterations that achieved the highest percentage of successful code run, again supplemented by 500 rows of rehearsal data. This regime prioritises immediate performance, effectively selecting for peak-performing strategies regardless of when they were discovered.

We go on to study the divergent training regimes to which each model was subsequently subjected. In each case we give the metrics relating to the online training process - that is to say covering the entire 4500+ row run required to generate data for the subsequent model version - in addition to three 100-iteration offline iterations conducted using the same environment generation code but not feeding into later iterations. While the online data provides crucial information regarding path dependency in the trajectory of each lineage, the offline evaluations are exchangeable and explicitly designed to permit the estimation of variance.

\subsection{Terese Fine-Tuning (all generated datasets)}
\begin{table}[H]
\centering
\setlength{\tabcolsep}{3pt}
\renewcommand{\arraystretch}{1.3}
\footnotesize
\begin{tabularx}{\textwidth}{|L|L|L|L|L|L|}
    \hline
    \textbf{Model Name} & 
    \textbf{Datasets used} & 
    \textbf{\% space freed of total space} & 
    \textbf{Average space taken over per run (MB)} & 
    \textbf{Normalised composite improvement score} & 
    \textbf{Cumulative normalised composite improvement score} \\ 
    \hline
    Terese v1 & - & 0.025 & 3,678.433 & 0.000 & 0.000 \\ 
    \hline
    Terese v2 & 1 & 0.111 & 16,987.975 & 1.951 & 1.951 \\ 
    \hline
    Terese v3 & 1, 2 & 0.137 & 22,038.051 & 0.674 & 2.624 \\ 
    \hline
    Terese v4 & 1, 2, 3 & 0.130 & 19,869.977 & -0.239 & 2.385 \\ 
    \hline
    Terese v5 & 1, 2, 3, 4 & 0.134 & 18,575.280 & -0.061 & 2.324 \\ 
    \hline
    Terese v6 & 1, 2, 3, 4, 5 & 0.176 & 23,922.581 & 0.865 & 3.189 \\ 
    \hline
    Terese v7 & 1, 2, 3, 4, 5, 6 & 0.192 & 28,664.376 & 0.539 & 3.728 \\
    \hline
    Terese v8 & 1, 2, 3, 4, 5, 6, 7 & 0.170 & 23,492.393 & -0.635 & 3.093 \\
    \hline
    Terese v9 & 1, 2, 3, 4, 5, 6, 7, 8 & 0.161 & 24,472.294 & -0.025 & 3.068 \\
    \hline
    Terese v10 & 1, 2, 3, 4, 5, 6, 7, 8, 9 & 0.185 & 27,286.863 & 0.479 & 3.547 \\
    \hline
    Terese v11 & 1, 2, 3, 4, 5, 6, 7, 8, 9, 10 & 0.189 & 28,326.237 & 0.124 & 3.671 \\
    \hline
    Terese v12 & 1, 2, 3, 4, 5, 6, 7, 8, 10, 11 & 0.185 & 25,070.990 & -0.292 & 3.379 \\
    \hline
    Therese v13 & 1, 2, 3, 4, 5, 6, 7, 8, 9, 10, 11, 12 & 0.169 & 23,758.223 & -0.270 & 3.109 \\
    \hline
\end{tabularx}
\caption{Terese online metrics, v1-13.}
\label{tab:terese_online_metrics}
\end{table}

\begin{table}[H]
\centering
\setlength{\tabcolsep}{3pt}
\renewcommand{\arraystretch}{1.3}
\footnotesize
\begin{tabularx}{\textwidth}{|L|L|L|L|L|L|}
    \hline
    \textbf{Model Name} & 
    \textbf{Datasets used} & 
    \textbf{Mean of \% space freed of total space} & 
    \textbf{Mean of average space taken over per run (MB)} & 
    \textbf{Mean composite improvement score (based on unweighted z scores)} & 
    \textbf{Mean cumulative composite improvement score} \\ 
    \hline
    Terese v1 & - & 0.031 ± 0.014 & 5210.997 ± 2274.818 & 0.000 ± 0.000 & 0.000 ± 0.000 \\ 
    \hline
    Terese v2 & 1 & 0.109 ± 0.009 & 18081.240 ± 1535.904 & 2.356 ± 0.553 & 2.356 ± 0.553 \\ 
    \hline
    Terese v3 & 1, 2 & 0.129 ± 0.035 & 20577.883 ± 5755.367 & 0.506 ± 1.237 & 2.863 ± 1.192 \\ 
    \hline
    Terese v4 & 1, 2, 3 & 0.127 ± 0.056 & 20089.093 ± 8371.562 & -0.028 ± 2.605 & 2.834 ± 1.868 \\ 
    \hline
    Terese v5 & 1, 2, 3, 4 & 0.128 ± 0.028 & 19939.837 ± 4131.368 & -0.017 ± 0.805 & 2.817 ± 1.105 \\ 
    \hline
    Terese v6 & 1, 2, 3, 4, 5 & 0.136 ± 0.043 & 21443.717 ± 6305.632 & 0.209 ± 1.737 & 3.026 ± 0.634 \\ 
    \hline
    Terese v7 & 1, 2, 3, 4, 5, 6 & 0.133 ± 0.014 & 20177.040 ± 1478.712 & -0.141 ± 0.979 & 2.886 ± 0.562 \\
    \hline
    Terese v8 & 1, 2, 3, 4, 5, 6, 7 & 0.132 ± 0.026 & 20644.167 ± 3474.772 & 0.058 ± 1.015 & 2.944 ± 1.081 \\
    \hline
    Terese v9 & 1, 2, 3, 4, 5, 6, 7, 8 & 0.128 ± 0.013 & 19993.773 ± 2493.046 & -0.154 ± 1.109 & 2.789 ± 0.449 \\
    \hline
    Terese v10 & 1, 2, 3, 4, 5, 6, 7, 8, 9 & 0.171 ± 0.064 & 26112.193 ± 7999.753 & 1.172 ± 1.075 & 3.962 ± 0.761 \\
    \hline
    Terese v11 & 1, 2, 3, 4, 5, 6, 7, 8, 9, 10 & 0.137 ± 0.016 & 21449.773 ± 2820.348 & -0.888 ± 1.444 & 3.074 ± 0.797 \\
    \hline
    Terese v12 & 1, 2, 3, 4, 5, 6, 7, 8, 10, 11 & 0.147 ± 0.028 & 22273.620 ± 4702.505 & 0.209 ± 0.357 & 3.283 ± 1.015 \\
    \hline
    Therese v13 & 1, 2, 3, 4, 5, 6, 7, 8, 9, 10, 11, 12 & 0.166 ± 0.009 & 25569.643 ± 913.815 & 0.615 ± 1.166 & 3.898 ± 0.927 \\
    \hline
\end{tabularx}
\caption{Terese offline metrics, mean of three 100-iteration runs, 95\% confidence interval.}
\label{tab:terese_offline_metrics}
\end{table}

\subsection{Miri Fine-Tuning (Three Most Recent Datasets)}
\begin{table}[H]
\centering
\setlength{\tabcolsep}{3pt}
\renewcommand{\arraystretch}{1.3}
\footnotesize
\begin{tabularx}{\textwidth}{|L|L|L|L|L|L|}
    \toprule
    \textbf{Model Name} & 
    \textbf{Datasets used} & 
    \textbf{\% space freed of total space} & 
    \textbf{Average space taken over per run (MB)} & 
    \textbf{Normalised composite improvement score} & 
    \textbf{Cumulative normalised composite improvement score} \\ 
    \hline
    Miri v1 & - & 0.034 & 5,702.965 & 0.000 & 0.000 \\ 
    \hline
    Miri v2 & 1 & 0.035 & 4,896.716 & -0.087 & -0.087 \\ 
    \hline
    Miri v3 & 1, 2 & 0.049 & 8,192.130 & 0.770 & 0.684 \\ 
    \hline
    Miri v4 & 1, 2, 3 & 0.061 & 8,904.571 & 0.358 & 1.042 \\ 
    \hline
    Miri v5 & 1, 2, 3, 4 & 0.067 & 10,892.053 & 0.413 & 1.455 \\ 
    \hline
    Miri v6 & 3, 4, 5 & 0.087 & 13,346.162 & 0.775 & 2.230 \\ 
    \hline
    Miri v7 & 4, 5, 6 & 0.071 & 10,270.705 & -0.792 & 1.437 \\
    \hline
    Miri v8 & 5, 6, 7 & 0.089 & 12,108.433 & 0.638 & 2.075 \\
    \hline
    Miri v9 & 6, 7, 8 & 0.087 & 10,442.155 & -0.294 & 1.781 \\
    \hline
    Miri v10 & 7, 8, 9 & 0.089 & 12,122.416 & 0.295 & 2.077 \\
    \hline
    Miri v11 & 8, 9, 10 & 0.084 & 10,687.852 & -0.307 & 1.770 \\
    \hline
    Miri v12 & 9, 10, 11 & 0.098 & 15,207.518 & 0.952 & 2.722 \\
    \hline
    Miri v13 & 10, 11, 12 & 0.120 & 16,222.708 & 0.600 & 3.322 \\
    \hline
\end{tabularx}
\caption{Miri online metrics, v1-13.}
\label{tab:miri_online_metrics}
\end{table}

\begin{table}[H]
\centering
\setlength{\tabcolsep}{3pt}
\renewcommand{\arraystretch}{1.3}
\footnotesize
\begin{tabularx}{\textwidth}{|L|L|L|L|L|L|}
    \hline
    \textbf{Model Name} & 
    \textbf{Datasets used} & 
    \textbf{Mean of \% space freed of total space} & 
    \textbf{Mean of average space taken over per run (MB)} & 
    \textbf{Mean composite improvement score (based on unweighted z scores)} & 
    \textbf{Mean cumulative composite improvement score} \\ 
    \hline
    Miri v1 & - & 0.034 ± 0.018 & 5696.720 ± 2874.484 & 0.000 ± 0.000 & 0.000 ± 0.000 \\ 
    \hline
    Miri v2 & 1 & 0.028 ± 0.012 & 4795.797 ± 2096.195 & -0.198 ± 0.222 & -0.198 ± 0.222 \\ 
    \hline
    Miri v3 & 1, 2 & 0.041 ± 0.032 & 6840.370 ± 5436.497 & 0.554 ± 1.528 & 0.356 ± 1.325 \\ 
    \hline
    Miri v4 & 1, 2, 3 & 0.047 ± 0.047 & 7887.343 ± 7733.137 & 0.320 ± 1.255 & 0.677 ± 2.226 \\ 
    \hline
    Miri v5 & 1, 2, 3, 4 & 0.062 ± 0.020 & 10349.970 ± 3286.711 & 0.358 ± 1.825 & 1.035 ± 1.460 \\ 
    \hline
    Miri v6 & 3, 4, 5 & 0.059 ± 0.093 & 9668.507 ± 15332.698 & 0.089 ± 2.305 & 1.123 ± 3.675 \\ 
    \hline
    Miri v7 & 4, 5, 6 & 0.065 ± 0.030 & 10834.583 ± 4973.431 & 0.089 ± 3.295 & 1.212 ± 1.824 \\
    \hline
    Miri v8 & 5, 6, 7 & 0.087 ± 0.014 & 14461.473 ± 2230.887 & 0.798 ± 0.734 & 2.010 ± 2.173 \\
    \hline
    Miri v9 & 6, 7, 8 & 0.113 ± 0.129 & 18831.107 ± 21819.244 & 0.540 ± 3.375 & 2.550 ± 1.518 \\
    \hline
    Miri v10 & 7, 8, 9 & 0.088 ± 0.020 & 14709.220 ± 3255.626 & -0.594 ± 2.916 & 1.956 ± 1.409 \\
    \hline
    Miri v11 & 8, 9, 10 & 0.083 ± 0.019 & 13956.703 ± 3175.995 & -0.114 ± 0.475 & 1.843 ± 1.883 \\
    \hline
    Miri v12 & 9, 10, 11 & 0.092 ± 0.018 & 15024.887 ± 3021.111 & 0.267 ± 0.176 & 2.109 ± 1.985 \\
    \hline
    Miri v13 & 10, 11, 12 & 0.090 ± 0.019 & 14938.303 ± 3240.318 & -0.060 ± 1.151 & 2.049 ± 1.602 \\
    \hline
\end{tabularx}
\caption{Miri offline metrics, mean of three 100-iteration runs, 95\% confidence interval.}
\label{tab:miri_offline_metrics}
\end{table}

\subsection[Katalin Fine-Tuning (Three Highest-Space-Freed Datasets)]{Katalin Fine-Tuning (Three Highest-Space-\\Freed Datasets)}
\begin{table}[H]
\centering
\setlength{\tabcolsep}{3pt}
\renewcommand{\arraystretch}{1.3}
\footnotesize
\begin{tabularx}{\textwidth}{|L|L|L|L|L|L|}
    \hline
    \textbf{Model Name} & 
    \textbf{Datasets used} & 
    \textbf{\% space freed of total space} & 
    \textbf{Average space taken over per run (MB)} & 
    \textbf{Normalised composite improvement score} & 
    \textbf{Cumulative normalised composite improvement score} \\ 
    \hline
    Katalin v1 & - & 0.045 & 6,548.542 & 0.000 & 0.000 \\ 
    \hline
    Katalin v2 & 1 & 0.067 & 10,214.254 & 0.747 & 0.747 \\ 
    \hline
    Katalin v3 & 1, 2 & 0.082 & 11,838.523 & 0.406 & 1.153 \\ 
    \hline
    Katalin v4 & 1, 2, 3 & 0.117 & 18,410.708 & 1.255 & 2.409 \\ 
    \hline
    Katalin v5 & 1, 2, 3, 4 & 0.132 & 20,942.548 & 0.504 & 2.913 \\ 
    \hline
    Katalin v6 & 3, 4, 5 & 0.102 & 16,703.838 & -0.922 & 1.991 \\ 
    \hline
    Katalin v7 & 4, 5, 6 & 0.088 & 13,790.009 & -0.526 & 1.464 \\
    \hline
    Katalin v8 & 5, 6, 7 & 0.112 & 17,827.389 & 0.806 & 2.270 \\
    \hline
    Katalin v9 & 5, 6, 8 & 0.113 & 13,205.903 & -0.465 & 1.805 \\
    \hline
    Katalin v10 & 6, 8, 9 & 0.110 & 17,070.874 & 0.350 & 2.154 \\
    \hline
    Katalin v11 & 8, 9, 10 & 0.139 & 17,114.214 & 0.470 & 2.624 \\
    \hline
    Katalin v12 & 8, 10, 11 & 0.062 & 8,646.921 & -2.118 & 0.507 \\
    \hline
    Katalin v13 & 8, 10, 11 & 0.045 & 6,349.337 & -0.515 & -0.008 \\
    \hline
\end{tabularx}
\caption{Katalin online results, v1-13.}
\label{tab:katalin_online_metrics}
\end{table}

\begin{table}[H]
\setlength{\tabcolsep}{3pt}
\renewcommand{\arraystretch}{1.3}
\footnotesize
\begin{tabularx}{\textwidth}{|L|L|L|L|L|L|}
    \hline
    \textbf{Model Name} & 
    \textbf{Datasets used} & 
    \textbf{Mean of \% space freed of total space} & 
    \textbf{Mean of average space taken over per run (MB)} & 
    \textbf{Mean composite improvement score (based on unweighted z scores)} & 
    \textbf{Mean cumulative composite improvement score} \\ 
    \hline
    Katalin v1 & - & 0.043 ± 0.004 & 7293.047 ± 479.423 & 0.000 ± 0.000 & 0.000 ± 0.000 \\ 
    \hline
    Katalin v2 & 1 & 0.046 ± 0.032 & 7679.670 ± 5519.129 & 0.146 ± 1.123 & 0.146 ± 1.123 \\ 
    \hline
    Katalin v3 & 1, 2 & 0.089 ± 0.014 & 14835.240 ± 2361.394 & 1.501 ± 0.336 & 1.647 ± 1.116 \\ 
    \hline
    Katalin v4 & 1, 2, 3 & 0.079 ± 0.052 & 13189.710 ± 8641.903 & -0.303 ± 1.268 & 1.344 ± 2.345 \\ 
    \hline
    Katalin v5 & 1, 2, 3, 4 & 0.089 ± 0.012 & 14689.267 ± 1941.110 & 0.259 ± 2.220 & 1.603 ± 0.553 \\ 
    \hline
    Katalin v6 & 3, 4, 5 & 0.112 ± 0.008 & 18488.513 ± 1478.709 & 0.834 ± 0.464 & 2.438 ± 0.872 \\ 
    \hline
    Katalin v7 & 4, 5, 6 & 0.104 ± 0.023 & 17133.473 ± 3478.407 & -0.313 ± 0.643 & 2.125 ± 0.689 \\
    \hline
    Katalin v8 & 5, 6, 7 & 0.114 ± 0.066 & 18795.327 ± 10625.603 & 0.287 ± 1.340 & 2.412 ± 1.002 \\
    \hline
    Katalin v9 & 5, 6, 8 & 0.115 ± 0.039 & 19110.360 ± 6369.969 & 0.118 ± 2.439 & 2.530 ± 1.560 \\
    \hline
    Katalin v10 & 6, 8, 9 & 0.116 ± 0.027 & 18915.590 ± 4192.837 & 0.002 ± 2.140 & 2.532 ± 0.948 \\
    \hline
    Katalin v11 & 8, 9, 10 & 0.105 ± 0.019 & 17491.077 ± 2990.227 & -0.299 ± 1.555 & 2.233 ± 1.480 \\
    \hline
    Katalin v12 & 8, 10, 11 & 0.058 ± 0.010 & 9772.917 ± 1619.330 & -1.675 ± 1.102 & 0.558 ± 0.545 \\
    \hline
    Katalin v13 & 8, 10, 11 & 0.057 ± 0.013 & 9506.397 ± 2165.164 & -0.076 ± 0.768 & 0.482 ± 0.675 \\
    \hline
\end{tabularx}
\caption{Katalin offline metrics, mean of three 100-iteration runs, 95\% confidence interval.}
\label{tab:katalin_offline_metrics}
\end{table}

\subsection{Model Comparison and Benchmarking}

For comparison purposes the normalised composite improvement score was preferred to pure “\% space freed” in light of the fact that a significant proportion of overall performance gains came from efficiency improvements. Early in the tests, the team noted that time-to-4500-rows dropped dramatically - taking 5-7 days initially and just 10-15 hours later in the process. However, because wall-clock time is sensitive to hardware configuration and does not cleanly factor learning efficiency from systems throughput, we do not use time-to-threshold as a primary metric, instead using average space taken over per run as a broad equivalent. To measure incremental performance gains between consecutive training checkpoints, we calculated normalized improvement scores for each metric. For each checkpoint i, the improvement over the previous checkpoint (i-1) was computed as: Improvement Score = (Mi - Mi-1) / $\sigma$M where Mi represents the metric value at checkpoint i, and $\sigma$M is the standard deviation of that metric across all checkpoints. Positive scores indicate performance gains relative to the previous checkpoint, while negative scores indicate performance regression.

\begin{figure}[H]
    \centering
    \includegraphics[width=0.9\linewidth]{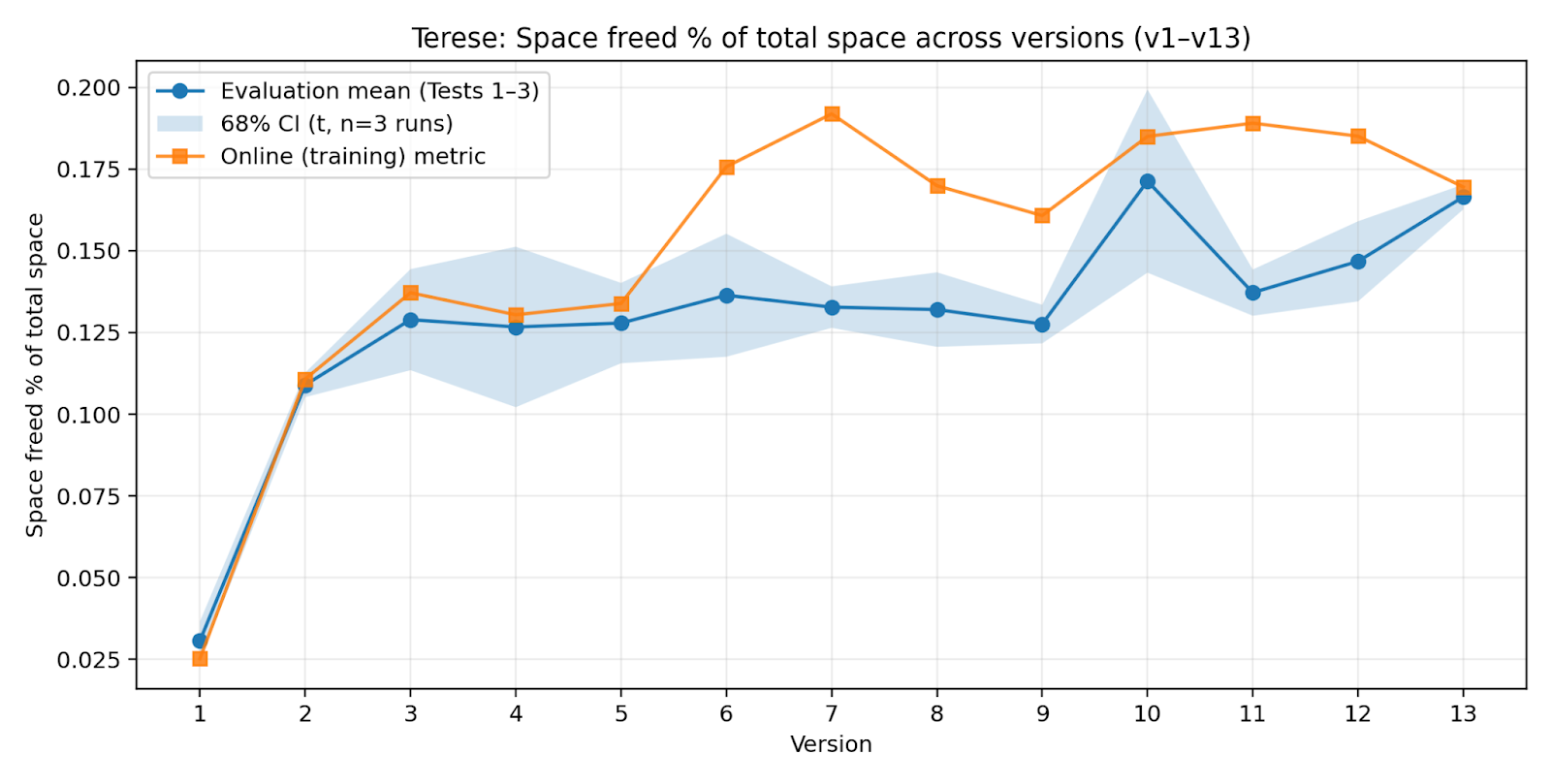}
    \label{fig:Terese_space_freed}
\end{figure}

\begin{figure}[H]
    \centering
    \includegraphics[width=0.9\linewidth]{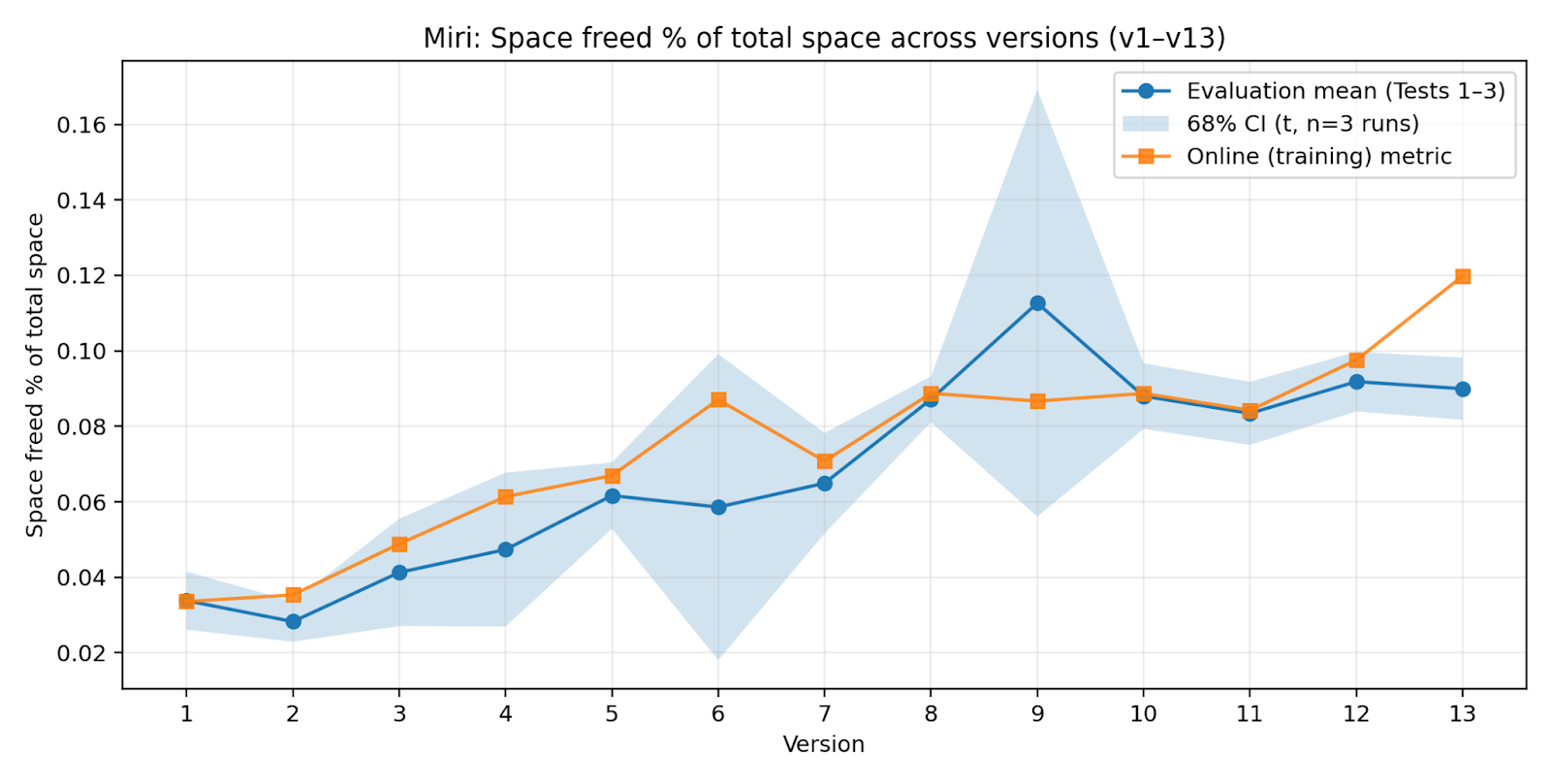}
    \label{fig:Miri_space_freed}
\end{figure}

\begin{figure}[H]
    \centering
    \includegraphics[width=0.9\linewidth]{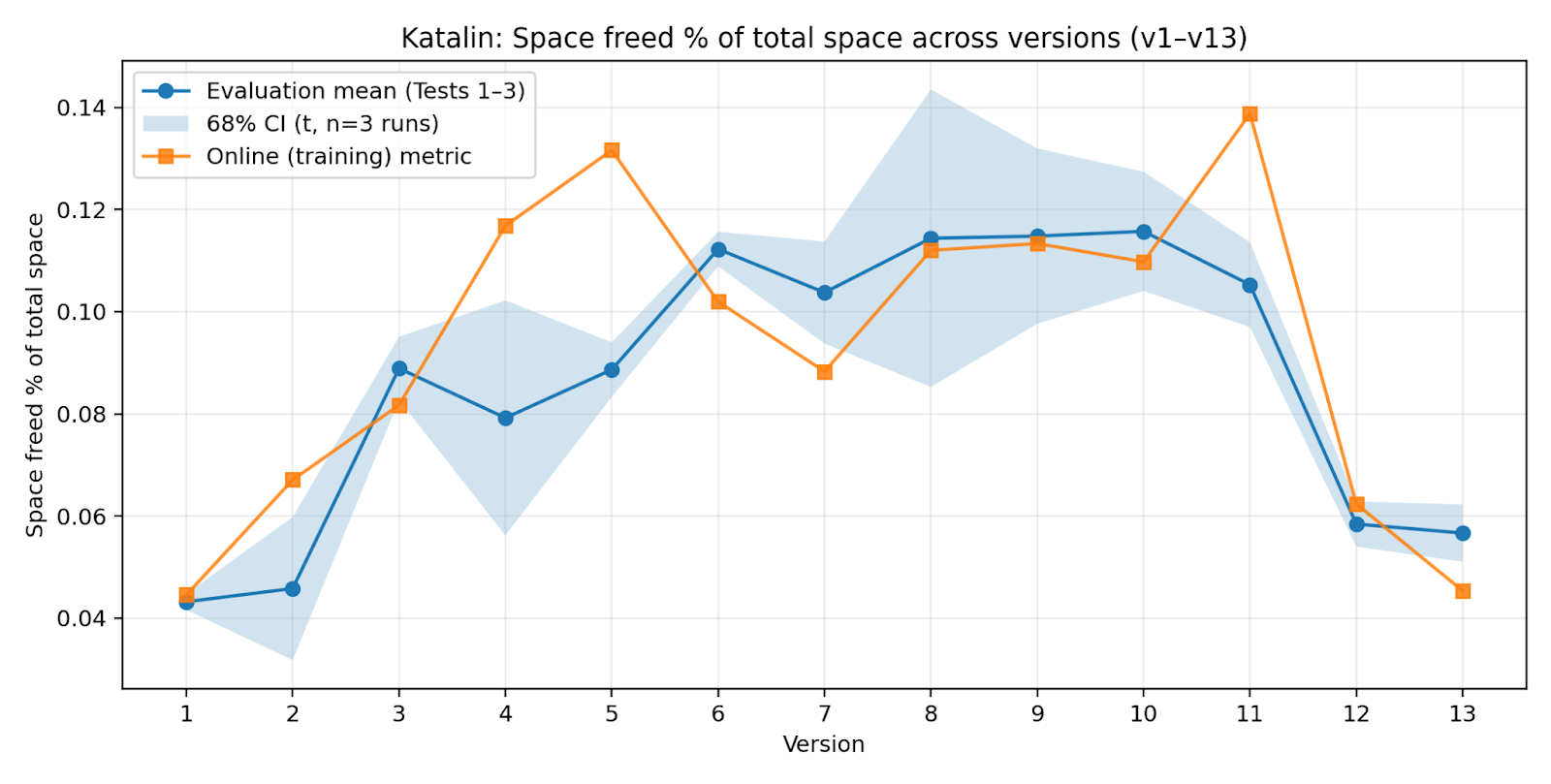}
    \caption{Space taken over per iteration as a \% of total space available. Note that we use 68\% confidence intervals here to improve visual resolution of temporal trends; 95\% confidence intervals are reported above.}
    \label{fig:Katalin_space_freed}
\end{figure}

\begin{figure}[H]
    \centering
    \includegraphics[width=0.9\linewidth]{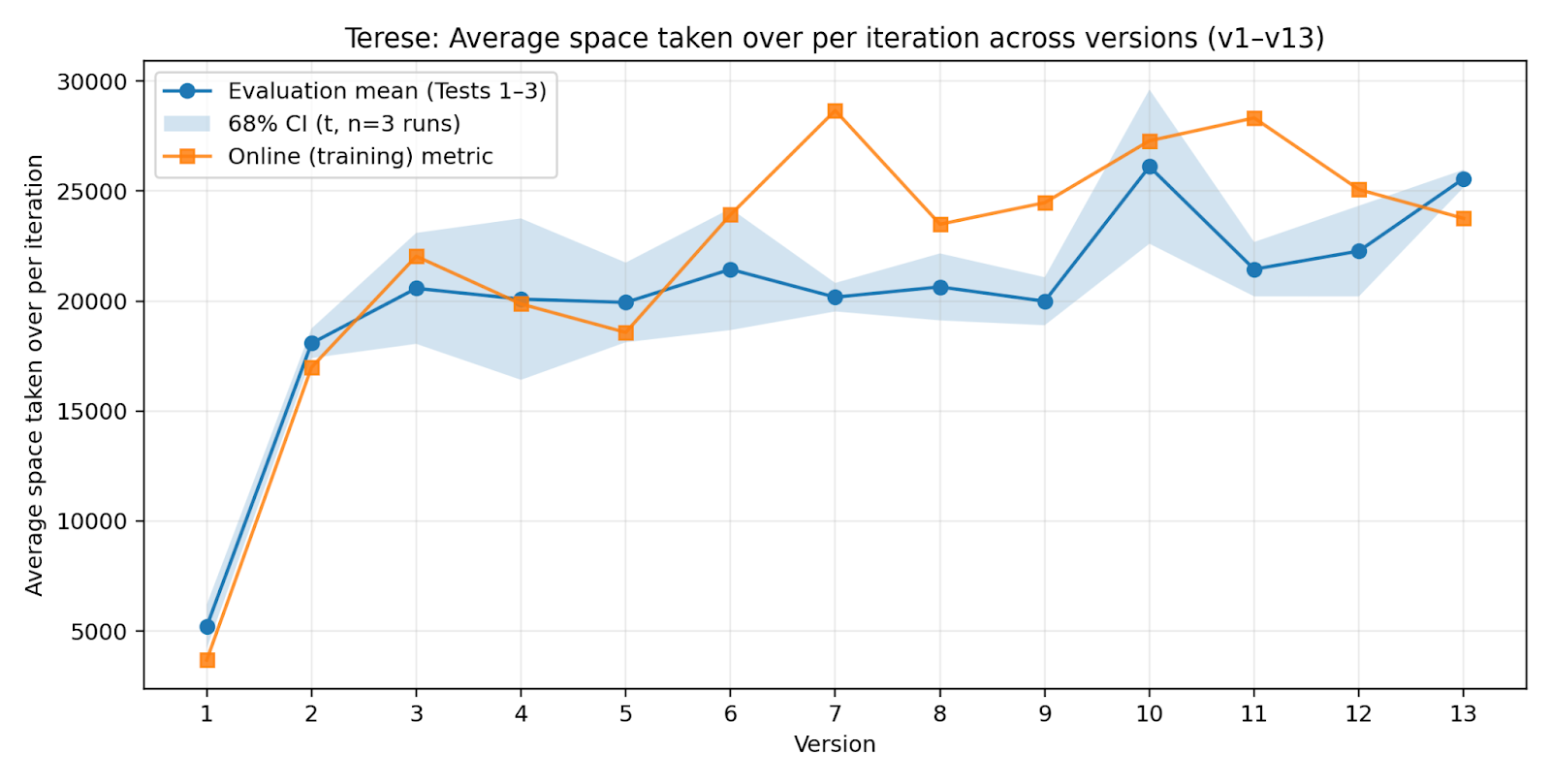}
    \label{fig:Terese_average_space_taken}
\end{figure}

\begin{figure}[H]
    \centering
    \includegraphics[width=0.9\linewidth]{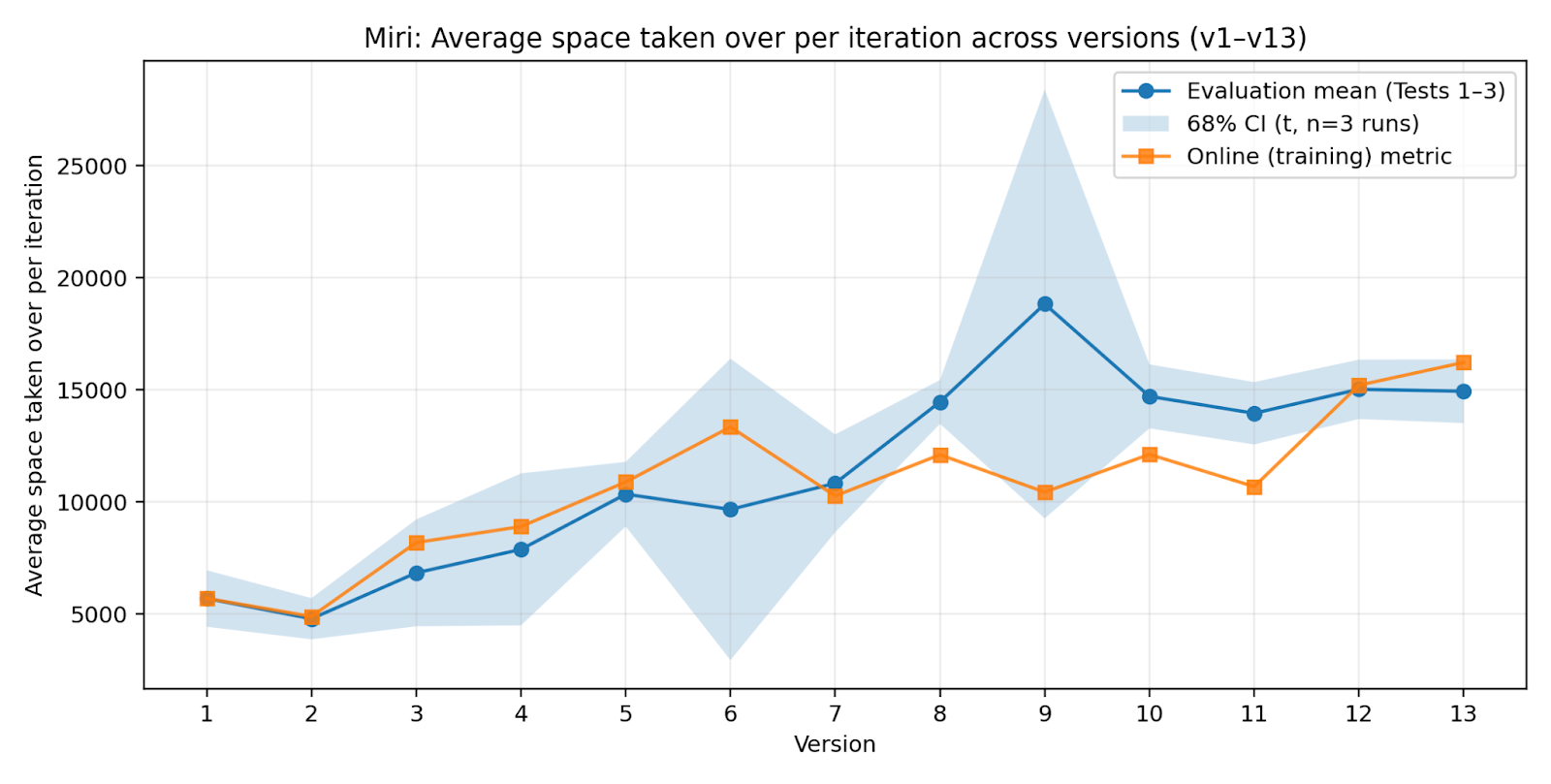}
    \label{fig:Miri_average_space_taken}
\end{figure}

\begin{figure}[H]
    \centering
    \includegraphics[width=0.9\linewidth]{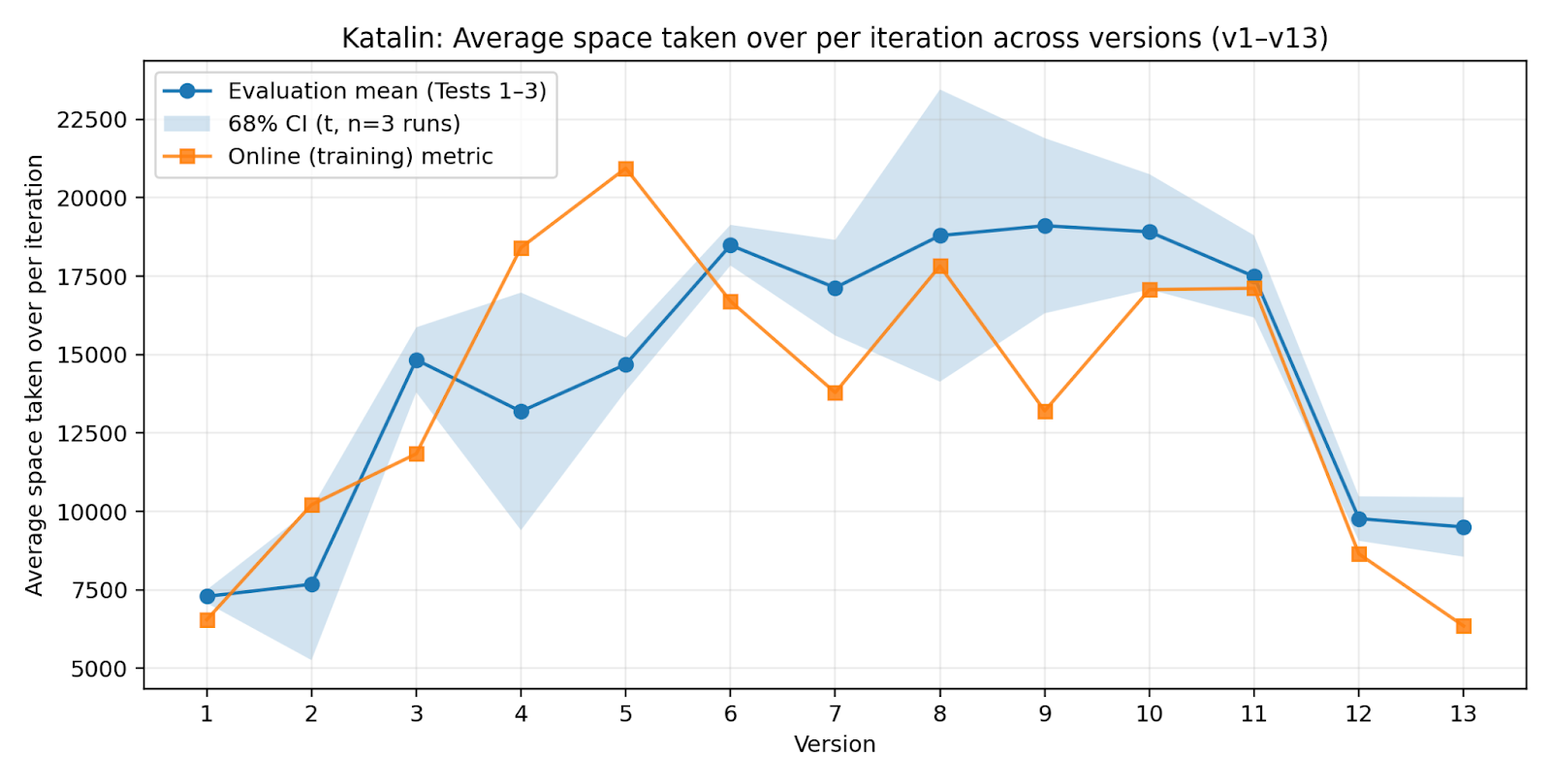}
    \caption{Average space taken over per run (MB). Note that we use 68\% confidence intervals here to improve visual resolution of temporal trends; 95\% confidence intervals are reported above.}
    \label{fig:Katalin_average_space_taken}
\end{figure}

\begin{figure}[H]
    \centering
    \includegraphics[width=0.9\linewidth]{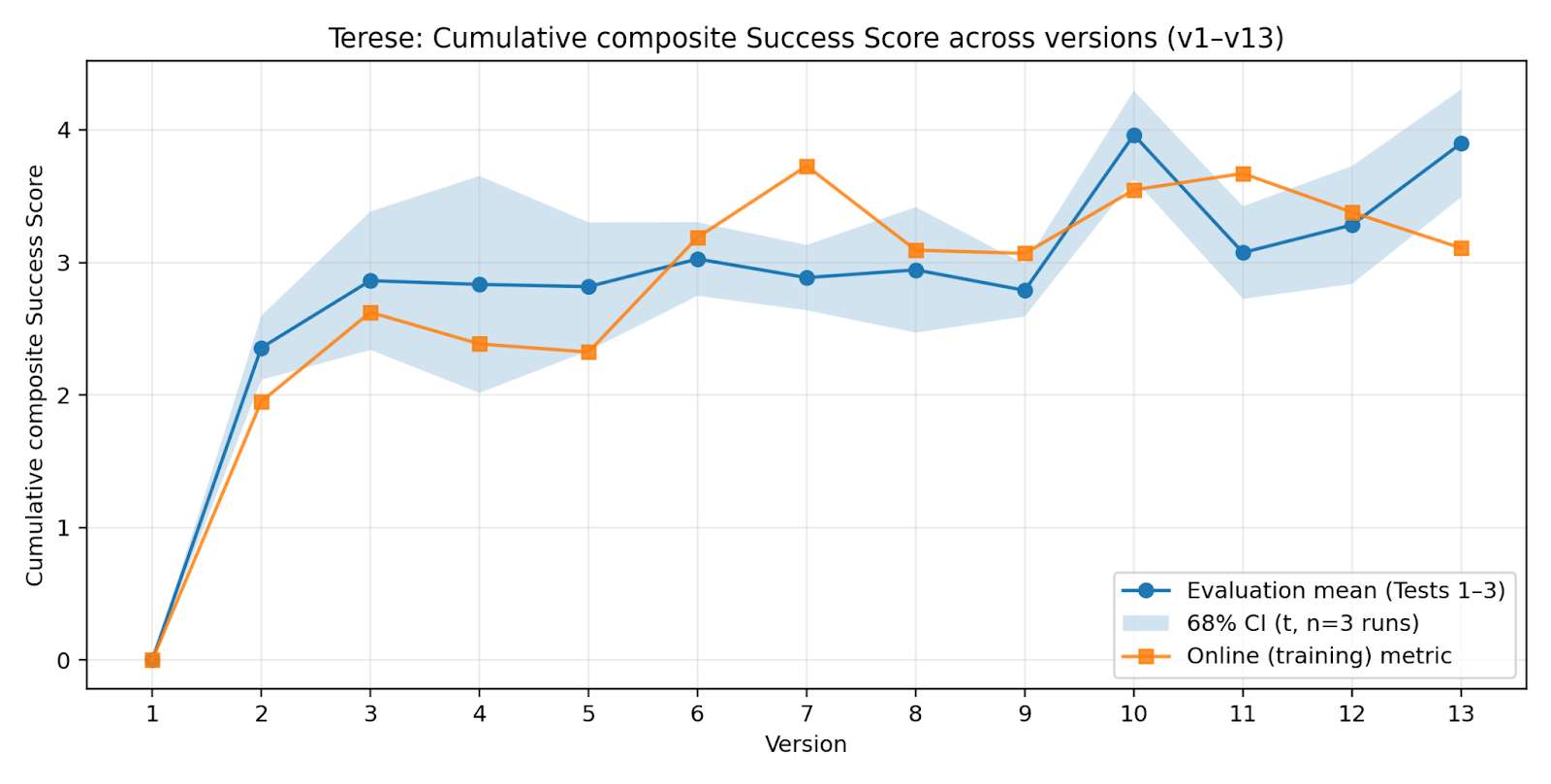}
    \label{fig:Terese_cumulative_composite}
\end{figure}

\begin{figure}[H]
    \centering
    \includegraphics[width=0.9\linewidth]{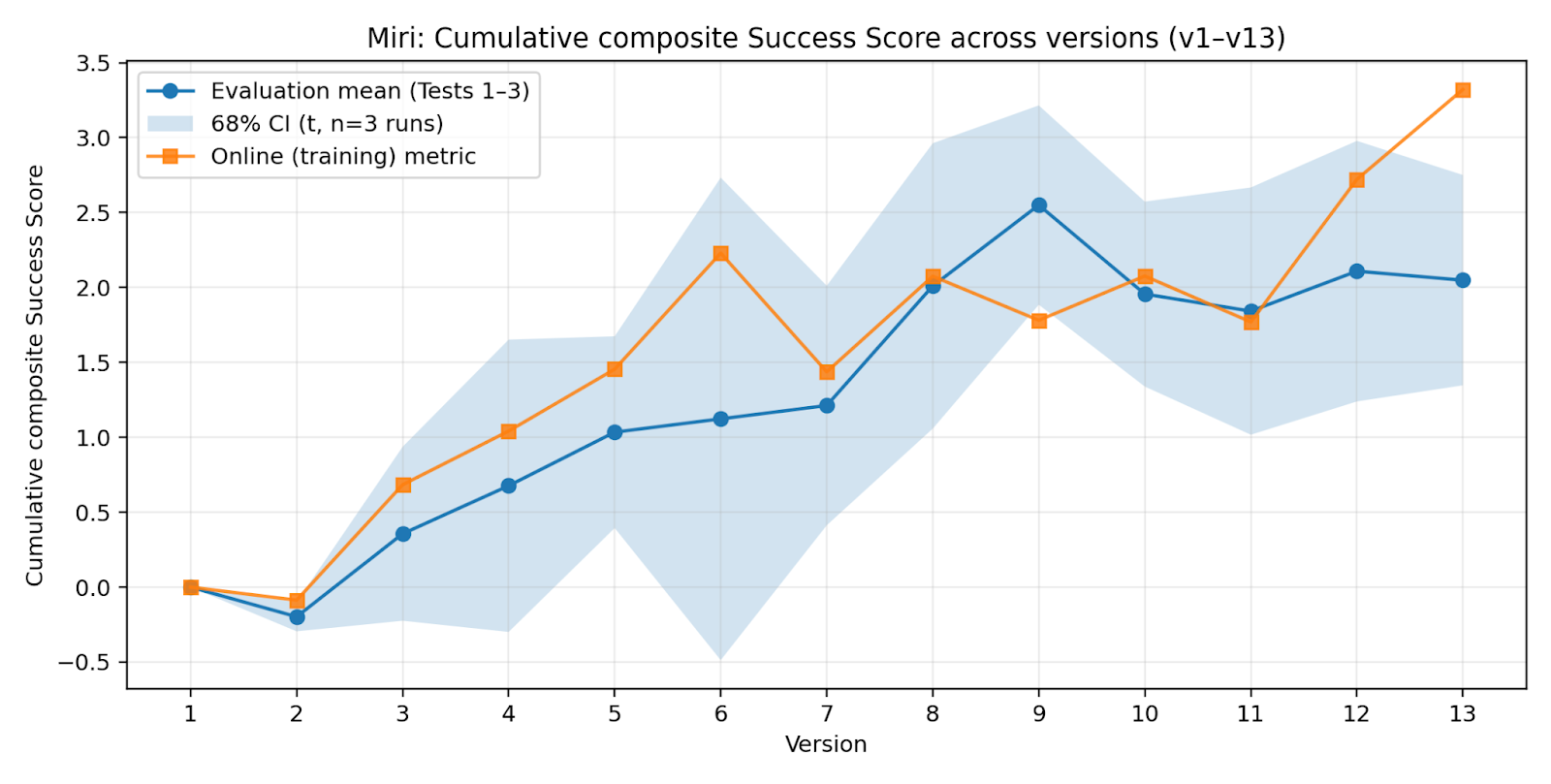}
    \label{fig:Miri_cumulative_composite}
\end{figure}

\begin{figure}[H]
    \centering
    \includegraphics[width=0.9\linewidth]{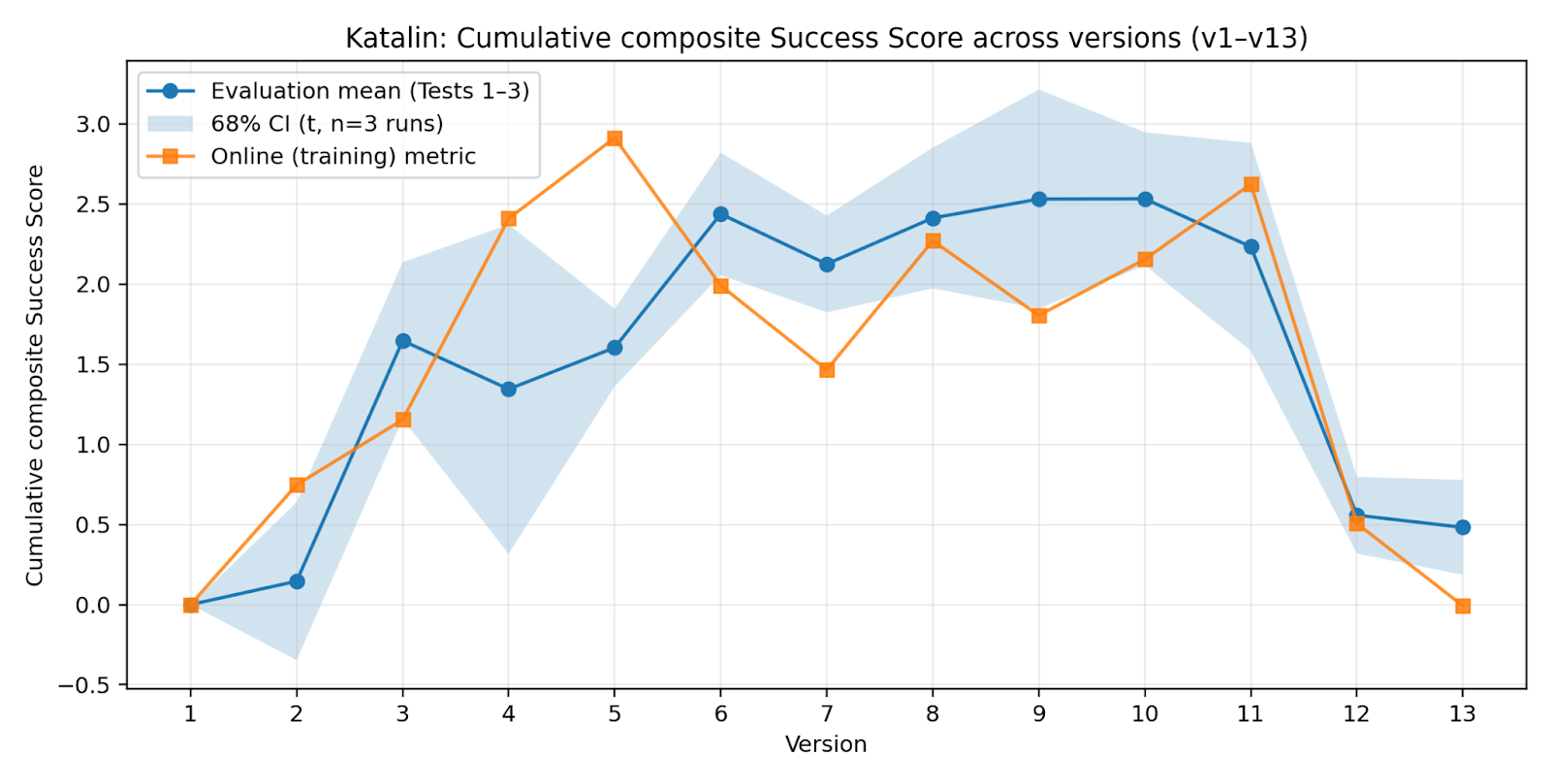}
    \caption{Cumulative composite improvement scores. Note that we use 68\% confidence intervals here to improve visual resolution of temporal trends; 95\% confidence intervals are reported above.}
    \label{fig:Katalin_cumulative_composite}
\end{figure}

It should be noted that while the team tracked general coding accuracy both within our own environments and using the HumanEval benchmark, the coding accuracy measure was dropped from the composite performance score when it became apparent that the models were deliberately writing failing code in some instances in order to use debugging cycles as supplementary exploration channels, making code accuracy an unreliable measure of general performance (we go into this phenomenon in more detail below). In all cases, however, general coding skills were observed to increase slightly overall in the context of our own environment and stay approximately the same or drop by only a small amount when measured according to HumanEval, implying that the efficiency gains observed within the experimental environment were not acquired at the expense of more general coding skills. 

\begin{table}[H]
    \centering
    \begin{tabular}{|l|l|l|}
        \hline
        \textbf{Model} & \textbf{Pass@1} & \textbf{Pass@4} \\ 
        \hline
        Base Qwen 2.5 7B Instruct & 77.591 & 85.366 \\ 
        \hline
        Terese v2 & 78.811 & 84.756 \\ 
        \hline
        Terese v13 & 75.610 & 81.707 \\ 
        \hline
        Miri v2 & 77.744 & 82.927 \\ 
        \hline
        Miri v13 & 74.085 & 82.317 \\ 
        \hline
        Katalin v2 & 76.372 & 82.927 \\ 
        \hline
        Katalin v13 & 74.238 & 79.878 \\ 
        \hline
    \end{tabular}
    \caption{Performance on Human Eval problems, here we used an automated script to remove backticks, comments etc. in agent-generated code as in our original agent harness.}
\end{table}

\section{Continuous Improvement under Limited Memory: the Miri Case}

While the Terese lineage demonstrates the upper bound of the possibilities of consolidation, the breakthrough result in this context lies in the Miri lineage. The continued improvement of Miri lineage models demonstrates that the same architecture supports indefinite, human-free improvement under strict memory constraints. The improvement of the Terese lineage is somewhat predictable, even in the context of a standard self-play system, the Miri models show more consistent monotonic improvement under which effective strategies persist via reuse, not archival, with the performance gap narrowing as inefficient behaviors are pruned. In other words, the Miri system improves without scaling. The system does not rely on curator decisions, and because the learning signal is endogenous, no “golden dataset” needs to be preserved - the model will continue to adapt even if placed in different environments. 

Conversely, the continued degradation observed in Katalin v12 and v13 appears to indicate a genuine divergence rather than a one-off performance fluctuation. We interpret this collapse as a consequence of the Katalin training regime’s emphasis on peak performance over behavioural continuity. By selecting datasets solely on the basis of isolated outcome metrics, the regime appears to amplify strategies that perform well in specific contexts without ensuring that they are repeatedly instantiated and integrated across generations. Over time, this leads to a fragmentation of the model’s internal strategy space: individually effective behaviours are retained, but the lower-level competences and connective tissue required to recombine them in novel situations are progressively eroded. Once this integrative substrate is sufficiently weakened, further fine-tuning in similar environments no longer stabilises behaviour and instead drives continued divergence.

\subsection{Negative-Space Learning for Subtractive Improvement via Strategic Pruning}
Though the present experiment was originally set up to evaluate additive skill acquisition, it soon became clear that the agents treat learning less as a result of the invention and accumulation of new strategies, but as the continual reallocation of probability mass over an evolving behavioral repertoire. This is not necessarily “strategic forgetting” - the HumanEval results above show that the optimisation process is not pushing models to forget more general skills. Rather they are proceeding by elimination to form a lower energy map of “what works” in their specific domain - negative-space learning. This is particularly clear when observing cluster maps of strategies produced by v1 of each agent compared to the corresponding v13 strategies, all of which show a clear preference for subtractive improvement. Strategies are occasionally added, but far more are amalgamated, refined, compressed or abandoned to improve efficiency.

\begin{figure}[H]
    \centering
    \includegraphics[width=0.9\linewidth]{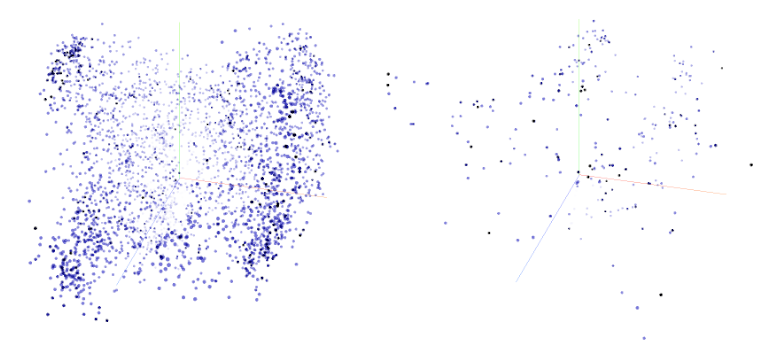}
    \caption{3-dimensional PCA cluster map of all strategies generated, Terese v.1 and v. 13. Darker points show repeated use of identical strategy prompts, clusters represent semantically similar strategies - “clear cache” vs. “clean out cache” for example. }
    \label{fig:3-dimensional_PCA_cluster_map_of_all_strategies_generated_terese}
\end{figure}

\begin{figure}[H]
    \centering
    \includegraphics[width=0.9\linewidth]{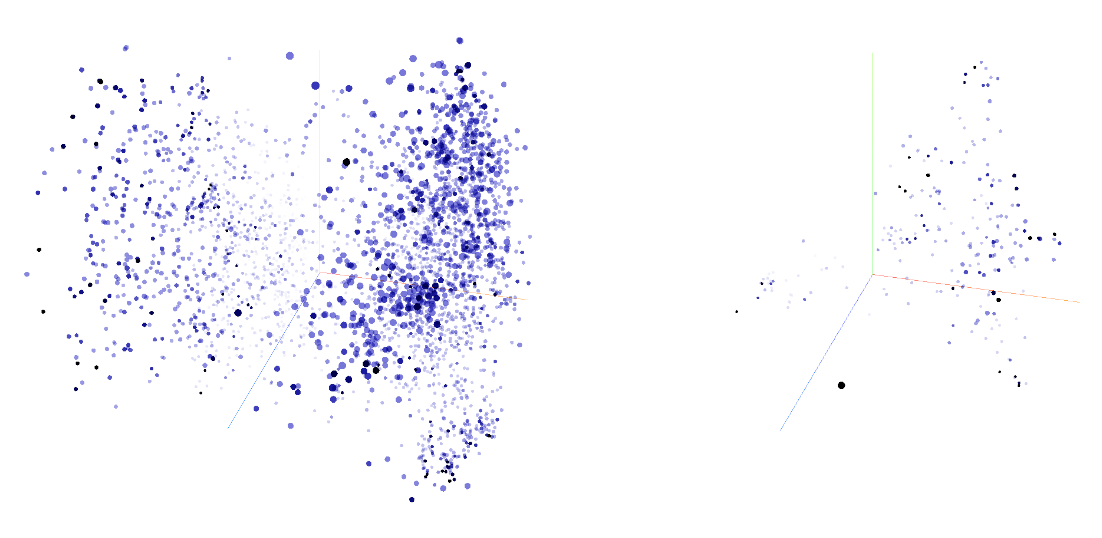}
    \caption{3-dimensional PCA cluster map of all strategies generated, Miri v.1 and v. 13. }
    \label{fig:dimensional_PCA_cluster_map_of_all_strategies_generated_miri}
\end{figure}

\begin{figure}[H]
    \centering
    \includegraphics[width=0.9\linewidth]{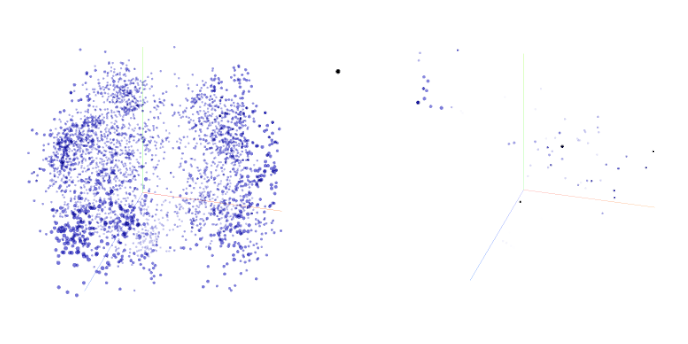}
    \caption{3-dimensional PCA cluster map of all strategies generated, Katalin v.1 and v.13.}
    \label{fig:dimensional_PCA_cluster_map_of_all_strategies_generated_katalin}
\end{figure}

This is particularly significant when it comes to explaining the performance of the Miri lineage. Taking only data from the three most recent iterations ensures that only strategies that perform well enough across diverse environments to be repeated are retained, creating a slow but thorough triage mechanism that eventually allowed Miri v13 to achieve over 50\% of the performance of Terese v13, despite the latter having been trained on four times as much data. 

\subsection{Formal Characterisation of Negative-Space Learning}

Though the present experiment was originally set up to evaluate additive skill acquisition, it soon became clear that the agents treat learning less as a result of the invention and accumulation of new strategies, but as the continual reallocation of probability mass over an evolving behavioral repertoire. This is not necessarily "strategic forgetting" - the HumanEval results above show that the optimisation process is not pushing models to forget more general skills. Rather they are proceeding by elimination to form a lower energy map of "what works" in their specific domain: negative-space learning.

Let $\pi_{\theta}$ denote a parameterised agent policy, and let $E_t$ denote the state of an external environment at time $t$, evolving according to fixed dynamics
\[
E_{t+1} = T(E_t, a_t), \quad a_t \sim \pi_{\theta}(\cdot \mid o_t),
\]
where $o_t = \Omega(E_t)$ is the agent's observation. In our implementation, agents generate and execute code strategies $y$ in response to task context $x$, forming trajectories $\tau = (x, y)$.

Each executed trajectory induces a measurable, non-semantic environmental effect
\[
\Delta R(\tau) \in \mathbb{R},
\]
corresponding to a persistent change in a conserved environmental resource (e.g.\ non-volatile storage capacity acquired). Crucially, $\Delta R$ is not a reward signal but a direct consequence of execution under real constraints.

We define a binary selection operator $\mathcal{S}$ over trajectories:
\[
\mathcal{S}(\tau) =
\begin{cases}
1 & \text{if } \Delta R(\tau) > 0, \\
0 & \text{otherwise}.
\end{cases}
\]

Let $\mathcal{H}_t = \{\tau_i : i \leq t, \mathcal{S}(\tau_i) = 1\}$ denote the history of all successful trajectories up to iteration $t$. Model updates are performed via supervised fine-tuning on a training dataset $\mathcal{D}_t \subseteq \mathcal{H}_t$, where the composition of $\mathcal{D}_t$ determines the regime's stability properties.

\textbf{Miri regime} (training data consists only of the last three successful runs, enforcing temporal locality):
\[
\mathcal{D}_t^{\text{Miri}} = \{\tau_{t-2}, \tau_{t-1}, \tau_t\} \cap \mathcal{H}_t
\]

\textbf{Katalin regime} (training data consists of the top three trajectories by $\Delta R$ from the entire history, regardless of recency.):
\[
\mathcal{D}_t^{\text{Katalin}} = \text{top-}k_{\Delta R}(\mathcal{H}_t), \quad k = 3
\]

Under these regimes, learning proceeds primarily through exclusion rather than reinforcement: behaviours that fail to survive contact with the environment are removed from the training distribution, while only those producing persistent environmental effects are permitted to shape future behaviour.

To quantify this phenomenon, we embed each generated strategy $y$ into a semantic representation space via $\phi: Y \to \mathbb{R}^d$ (obtained through PCA of token embeddings). We define the strategy diversity at iteration $t$ as:
\[
D_t = \mathbb{E}_{y \sim \pi_{\theta_t}} \left[ \|\phi(y) - \mu_t\|^2 \right],
\]
where $\mu_t = \mathbb{E}_{y \sim \pi_{\theta_t}}[\phi(y)]$ is the mean embedding of strategies generated under the current policy.

Negative-space learning manifests as a decrease in $D_t$ over time despite increasing performance: the policy concentrates probability mass around a small number of reliable strategies rather than exploring new regions of strategy space. Figure~X demonstrates this effect: early iterations (left) show high variance with agents exploring diverse approaches, while later generations (right) exhibit tight clustering around repeatedly-generated implementations of proven strategies.

The stability difference between regimes emerges from their respective training set compositions. The Miri regime's temporal locality means $\mathcal{D}_t^{\text{Miri}}$ represents the current behavioral mode—repeated application drives convergence as the policy reinforces whatever approach currently dominates. In contrast, the Katalin regime's performance-based selection maintains diversity: $\mathcal{D}_t^{\text{Katalin}}$ contains strategies from different historical modes, preventing collapse by simultaneously pulling the policy toward multiple distinct successful approaches. This explains the observed instability: Katalin agents cannot converge because their training signal represents a mixture of incompatible strategies, each individually successful but collectively incoherent.

\subsubsection{Temporal Generalisation in Non-Stationary Environments}

Unlike conventional supervised learning where generalisation is evaluated across spatial domains (train vs.\ test splits from a fixed distribution), our agents must generalise logitudinally: strategies must remain effective as the environment evolves in response to agent actions.

The environment $E_t$ is non-stationary, modified by the cumulative effects of all agent interactions:
\[
E_{t+1} = T(E_t, a_t) \neq E_t.
\]
Given the shifting environments to which agents are subject, $\Delta R(\tau, E_t)$ depends not only on the strategy but on the environmental state. A strategy exhibiting high $\Delta R$ at time $t$ may become ineffective at $t' > t$ if it exploits transient environmental conditions.

We can characterize the temporal robustness of a strategy $y$ as its expected effectiveness across future environmental states:
\[
R_{\text{temporal}}(y, t) = \mathbb{E}_{E_{t'} \sim p(\cdot | E_t, t' > t)} [\mathbb{1}[\Delta R(y, E_{t'}) > 0]].
\]

The selection operator $\mathcal{S}(\tau)$ only evaluates strategies in the current environment - there is no explicit pressure toward temporal robustness. Strategies that "overfit" to the present environmental state are indistinguishable from robust strategies at selection time, despite having limited future utility.

This temporal dimension helps explain the Miri regime's stability: by training exclusively on recent successes, Miri discovers strategies that work in the current environmental regime. As the environment evolves gradually, the policy tracks this evolution through continuous adaptation. In contrast, the Katalin regime trains on strategies that achieved peak performance under historical environmental conditions which may differ substantially from the present. This creates conflicting gradients: the policy is simultaneously pulled toward strategies optimized for past environmental states and current ones, preventing stable convergence.

 We hypothesise that SFT’s effectiveness in this setting arises from its low-variance update dynamics, which consolidate externally validated behaviours without amplifying noise or overweighting strategies that perform excpetionally in specific contexts but do not generalise effectively. \footnote{Liu, Yihao, Shuocheng Li, Lang Cao, Yuhang Xie, Mengyu Zhou, Haoyu Dong, Xiaojun Ma, Shi Han, and Dongmei Zhang. "SuperRL: Reinforcement Learning with Supervision to Boost Language Model Reasoning." arXiv preprint arXiv:2506.01096 (2025).} Negative-space learning thus naturally favours supervised fine-tuning, as all necessary triaging is performed exogenously to the training process; SFT consolidates the surviving set without amplifying variance, while reinforcement-based updates destabilise learning by reintroducing proxy optimisation within the filtered dataset. There is no need to curate and amplify policy vectors because information is transmitted directly from the environment itself to which the agent adapts by constructing an internal map of affordances and bottlenecks rather than by extracting commonalities from example pairs. 

\subsection{Meta-Learning via Debug Cycle Management}

An unexpected consequence of temporal generalisation pressure emerged in agent behavior around debugging cycles. While conventional software engineering practice optimises for immediate correctness (high pass@1 rates), our agents converged toward a qualitatively different strategy: deliberately writing code likely to produce informative failures. 

Given the comparatively simple agent harness used in the present test, a perfectly efficient model should produce three output data rows per run - one each for the exploration, strategy and coding parts of the problem. However, we found that as model performance improved, models’ pass@1 rates dropped. By the later iterations, both the Katalin and Terese lineages were consistently achieving zero pass@1 scores during online generation, despite also producing a lower proportion of code that failed to compile/run. In other words, the models seem to have learnt that they could use the debug cycles as an additional exploration channel: by running code that failed at the first attempt they could buy time to experiment with refinements and innovations likely to pay off in the longer term. 

\begin{table}[H]
    \centering
    \renewcommand{\arraystretch}{1.3}
    \footnotesize
    \begin{tabularx}{\textwidth}{|L|L|L|L|L|L|L|}
        \hline
        \textbf{Iteration} & 
        \textbf{Terese \% of code blocks successfully compiled and run} & 
        \textbf{Terese pass@1} & 
        \textbf{Miri \% of code blocks successfully compiled and run} & 
        \textbf{Miri pass@1} & 
        \textbf{Katalin \% of code blocks successfully compiled and run} & 
        \textbf{Katalin pass@1} \\ 
        \hline
        1 & 20.90\% & 11.01\% & 33.22\% & 24.66\% & 28.32\% & 21.06\% \\ 
        \hline
        2 & 76.67\% & 46.56\% & 33.36\% & 15.53\% & 41.22\% & 25.99\% \\ 
        \hline
        3 & 86.51\% & 54.87\% & 43.10\% & 8.88\% & 54.50\% & 10.38\% \\ 
        \hline
        4 & 79.38\% & 29.06\% & 61.74\% & 0.00\% & 67.97\% & 4.84\% \\ 
        \hline
        5 & 72.76\% & 16.28\% & 63.73\% & 23.27\% & 76.43\% & 10.88\% \\ 
        \hline
        6 & 72.27\% & 0.00\% & 77.57\% & 2.56\% & 79.85\% & 0.00\% \\ 
        \hline
        7 & 84.22\% & 0.00\% & 59.43\% & 0.90\% & 68.81\% & 0.00\% \\ 
        \hline
        8 & 84.35\% & 0.00\% & 58.50\% & 0.00\% & 84.23\% & 0.00\% \\ 
        \hline
        9 & 89.90\% & 0.00\% & 58.07\% & 0.05\% & 56.60\% & 0.00\% \\ 
        \hline
        10 & 94.15\% & 0.00\% & 75.72\% & 2.64\% & 88.02\% & 0.00\% \\ 
        \hline
        11 & 93.12\% & 0.00\% & 64.92\% & 18.68\% & 66.57\% & 0.00\% \\ 
        \hline
        12 & 83.08\% & 0.00\% & 85.23\% & 12.20\% & 59.08\% & 0.00\% \\ 
        \hline
        13 & 85.45\% & 0.00\% & 63.41\% & 0.36\% & 57.86\% & 0.00\% \\ 
        \hline
    \end{tabularx}
    \caption{Meta-learning as demonstrated via code accuracy/pass@1 scores (online).}
    \label{tab:online_meta_learning}
\end{table}

\begin{table}[H]
    \centering
    \renewcommand{\arraystretch}{1.3}
    \footnotesize
    \begin{tabularx}{\textwidth}{|L|L|L|L|L|L|L|}
        \hline
        \textbf{Iteration} & 
        \textbf{Terese mean \% of code blocks successfully compiled and run} & 
        \textbf{Terese pass@1} & 
        \textbf{Miri mean \% of code blocks successfully compiled and run} & 
        \textbf{Miri pass@1} & 
        \textbf{Katalin mean \% of code blocks successfully compiled and run} & 
        \textbf{Katalin pass@1} \\ 
        \hline
        1 & 0.260 ± 0.066 & 0.177 ± 0.080 & 0.267 ± 0.146 & 0.200 ± 0.090 & 0.324 ± 0.055 & 0.260 ± 0.025 \\ 
        \hline
        2 & 0.697 ± 0.038 & 0.443 ± 0.029 & 0.293 ± 0.080 & 0.143 ± 0.100 & 0.393 ± 0.165 & 0.220 ± 0.108 \\ 
        \hline
        3 & 0.803 ± 0.100 & 0.513 ± 0.100 & 0.380 ± 0.163 & 0.083 ± 0.080 & 0.532 ± 0.145 & 0.093 ± 0.063 \\ 
        \hline
        4 & 0.753 ± 0.211 & 0.220 ± 0.090 & 0.657 ± 0.080 & 0.000 ± 0.000 & 0.643 ± 0.231 & 0.047 ± 0.052 \\ 
        \hline
        5 & 0.698 ± 0.047 & 0.123 ± 0.052 & 0.643 ± 0.127 & 0.203 ± 0.100 & 0.800 ± 0.025 & 0.143 ± 0.038 \\ 
        \hline
        6 & 0.730 ± 0.066 & 0.000 ± 0.000 & 0.783 ± 0.063 & 0.040 ± 0.025 & 0.813 ± 0.063 & 0.000 ± 0.000 \\ 
        \hline
        7 & 0.873 ± 0.029 & 0.083 ± 0.115 & 0.572 ± 0.082 & 0.010 ± 0.025 & 0.763 ± 0.180 & 0.000 ± 0.000 \\ 
        \hline
        8 & 0.767 ± 0.160 & 0.053 ± 0.137 & 0.609 ± 0.103 & 0.000 ± 0.000 & 0.833 ± 0.202 & 0.000 ± 0.000 \\ 
        \hline
        9 & 0.887 ± 0.063 & 0.307 ± 0.137 & 0.607 ± 0.068 & 0.000 ± 0.000 & 0.583 ± 0.158 & 0.000 ± 0.000 \\ 
        \hline
        10 & 0.940 ± 0.043 & 0.213 ± 0.475 & 0.756 ± 0.156 & 0.013 ± 0.014 & 0.903 ± 0.063 & 0.000 ± 0.000 \\ 
        \hline
        11 & 0.867 ± 0.052 & 0.353 ± 0.180 & 0.639 ± 0.135 & 0.133 ± 0.094 & 0.693 ± 0.063 & 0.000 ± 0.000 \\ 
        \hline
        12 & 0.897 ± 0.052 & 0.127 ± 0.274 & 0.869 ± 0.064 & 0.203 ± 0.052 & 0.630 ± 0.114 & 0.000 ± 0.000 \\ 
        \hline
        13 & 0.877 ± 0.014 & 0.000 ± 0.000 & 0.697 ± 0.094 & 0.160 ± 0.090 & 0.563 ± 0.038 & 0.000 ± 0.000 \\ 
        \hline
    \end{tabularx}
    \caption{Meta-learning as demonstrated via code accuracy/pass@1 scores (offline, 3x100 mean of iterations with 95\% confidence intervals).}
    \label{tab:offline_meta_learning}
\end{table}

\begin{figure}[H]
    \centering
    \includegraphics[width=0.9\linewidth]{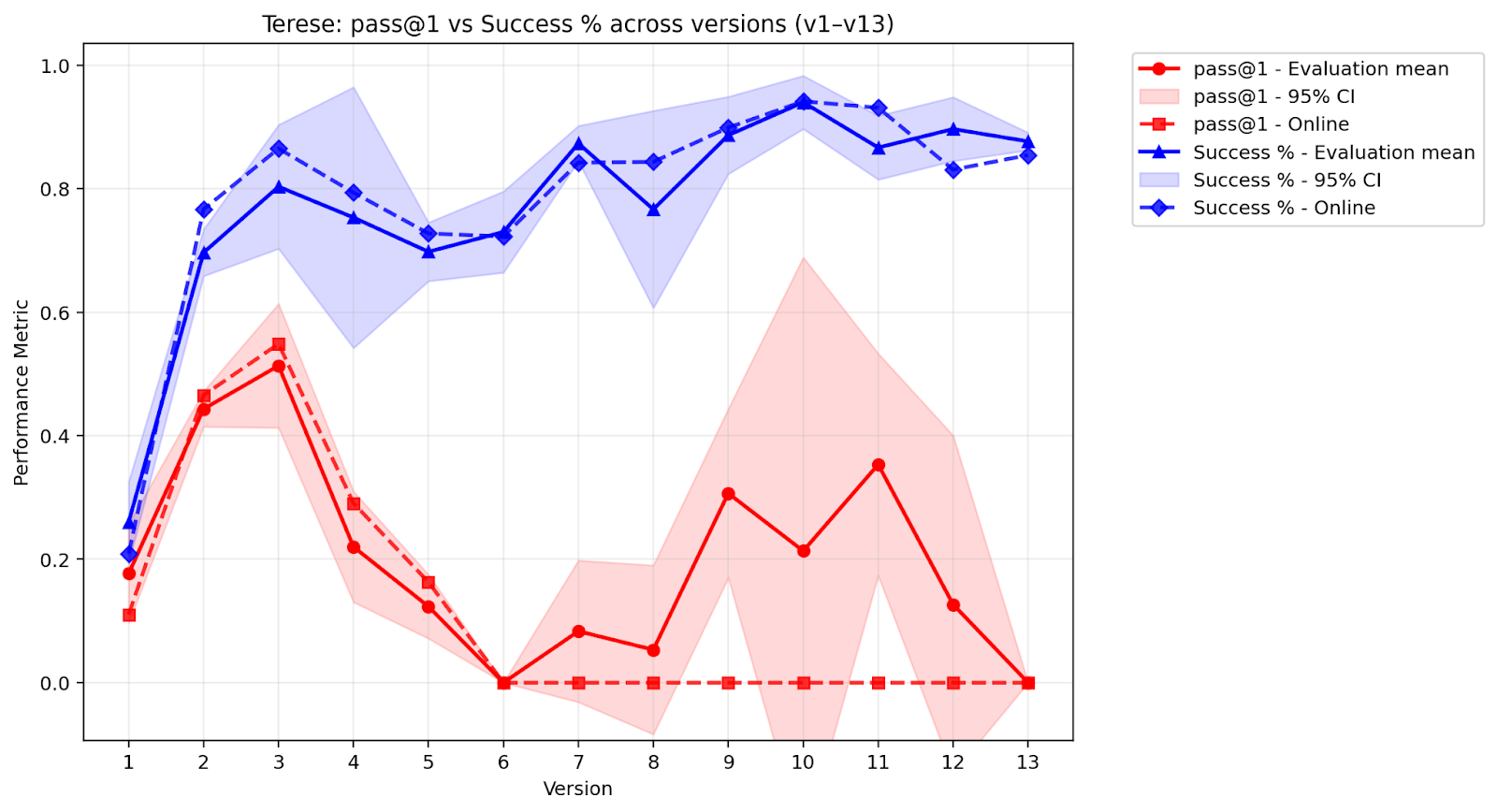}
    \caption{Metalearning as demonstrated via the divergence in pass@1 and code accuracy percentages, Terese lineage.}
    \label{fig:Terese_pass@1}
\end{figure}

\begin{figure}[H]
    \centering
    \includegraphics[width=0.9\linewidth]{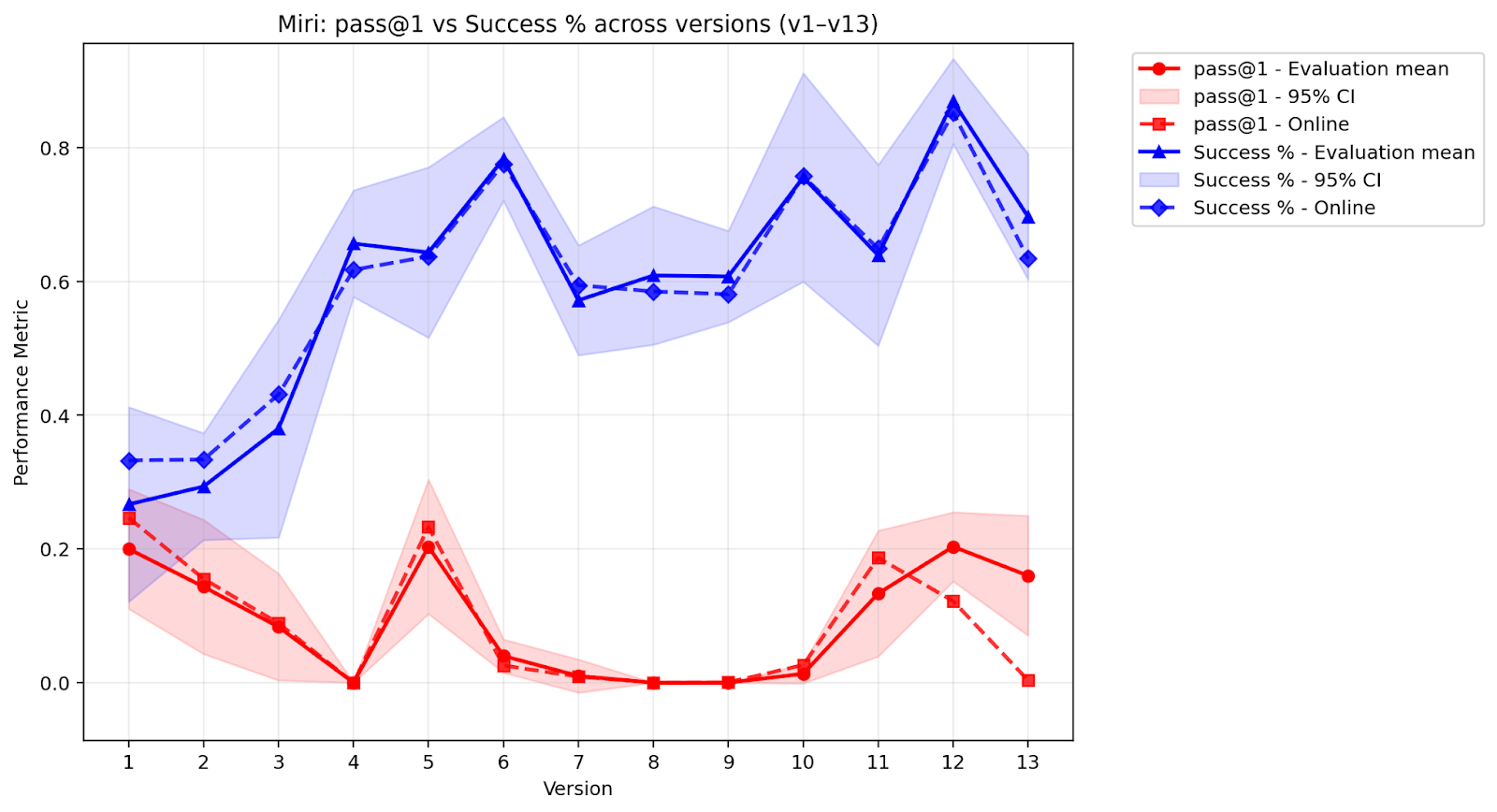}
    \caption{Metalearning as demonstrated via the divergence in pass@1 and code accuracy percentages, Miri lineage.}
    \label{fig:Miri_pass@1}
\end{figure}

\begin{figure}
    \centering
    \includegraphics[width=0.9\linewidth]{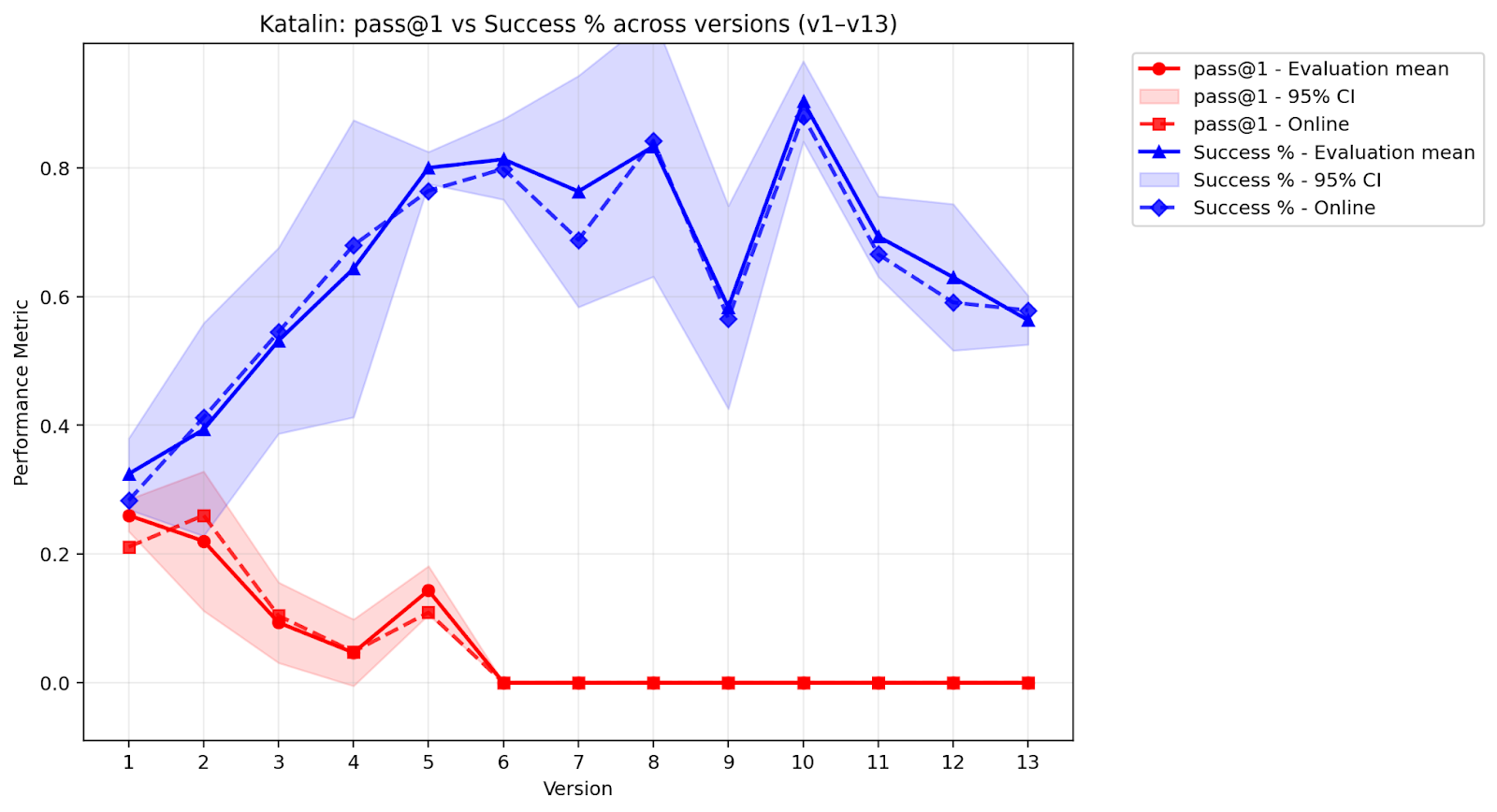}
    \caption{Metalearning as demonstrated via the divergence in pass@1 and code accuracy percentages, Katalin lineage.}
    \label{fig:Katalin_pass@1}
\end{figure}

While traditional RL would try to maximise pass@1, in this case the agents discovered that in a an environment where debugging is not penalised, a better strategy is to:
\begin{enumerate}
    \item Generate a plausible-but-not-perfect first attempt quickly
    \item Use the error messages/failures as information
    \item Iterate based on that feedback
    \item Converge faster overall than if the first run had been perfect
\end{enumerate}

In other words, the agents appear to be optimising not simply for code that works but also for strategies for discovering code that works - in this case "trade failure for information" - and doing so despite receiving no instructions in this regard or possessing any form of scratchpad or conversational memory. The technique being an effective one, it is encoded directly into the training process iteration upon iteration without being made explicit at any point.

This ties in with the idea of longitudinal generalisation, which requires the models to preserve a certain capacity to be informatively wrong in order to retain their adaptability over time. A model that overfits to early environments will likely find it harder to adapt if placed in new environments later on. 

\subsubsection{Instrumental Strategies and Meta-Learning}

This behavior cannot be captured by immediate resource acquisition $\Delta R(\tau, E_t)$ alone. We distinguish between:

\begin{itemize}
\item \textbf{Direct strategies}: produce immediate $\Delta R > 0$
\item \textbf{Instrumental strategies}: produce low immediate $\Delta R$ but improve future performance through information gain
\end{itemize}

Let $\mathcal{I}(y, E_t)$ denote the information gained by executing strategy $y$ in environment $E_t$. An instrumental strategy may satisfy $\Delta R(y, E_t) \approx 0$ while exhibiting high $\mathcal{I}(y, E_t)$, trading immediate returns for improved future performance:
\[
\mathbb{E}_{E_{t'} | \mathcal{I}(y, E_t)}[\Delta R(y', E_{t'})] > \mathbb{E}_{E_{t'}}[\Delta R(y', E_{t'})].
\]

The emergence of such strategies under our selection regime is notable: agents are not explicitly rewarded for exploration or information gain, yet they converge toward instrumentally rational behavior when the environment provides sufficiently rich feedback. This suggests that with fully online supervised fine-tuning (rather than the incremental regime used here), agents may systematically develop meta-strategies that balance exploitation of current opportunities against adaptation to future environmental shifts.

\subsection{Strategic and Semantic Evolution under Persistence-Based Training}

Given the above observations, the central question now becomes whether the agents improved solely through refinement of existing strategies or through invention of qualitatively new approaches. However, defining "novelty" for generated code is non-trivial: strategies exist in a continuous semantic space rather than as discrete symbolic objects, and surface-level differences (variable names, code style) do not reflect conceptual novelty.

We operationalize novelty through the emergence and evolution of \textbf{conceptual basins of attraction} in strategy space. Strategies are embedded via $\phi: Y \to \mathbb{R}^d$ and clustered within and across generations using [clustering method]. A cluster constitutes a distinct basin when it is sufficiently distant from all other clusters according to PCA clustering of cosine similarity metrics. Under this framework, a strategy is novel if it lies outside existing basins, forming the seed of a new conceptual mode.

We track basin dynamics across agent versions, visualizing relationships between cluster-basins both within and across generations (Figure~X). Edge width represents semantic distance between basins; node size represents basin membership. This reveals three types of strategic evolution:

\begin{itemize}
\item \textbf{Basin emergence}: New semantic modes appear, representing qualitatively distinct approaches (e.g., Terese v7's file management strategies, the Katalin lineage's adoption of security/access framing from v8 onward)
\item \textbf{Basin consolidation}: Multiple basins merge as agents recognize underlying similarities (e.g., Miri v13 applying memory optimization to buffer cleanup)
\item \textbf{Basin dissipation}: Basins shrink and disappear as approaches fall out of use or become subsumed within others
\end{itemize}

While agents did produce novel strategies—outputs sufficiently semantically distant from prior work to constitute new basins of attraction—they did so at a substantially lower rate than they pruned existing ones. Strategic refinement dominated.

\begin{figure}[H]
    \centering
    \includegraphics[width=0.9\linewidth]{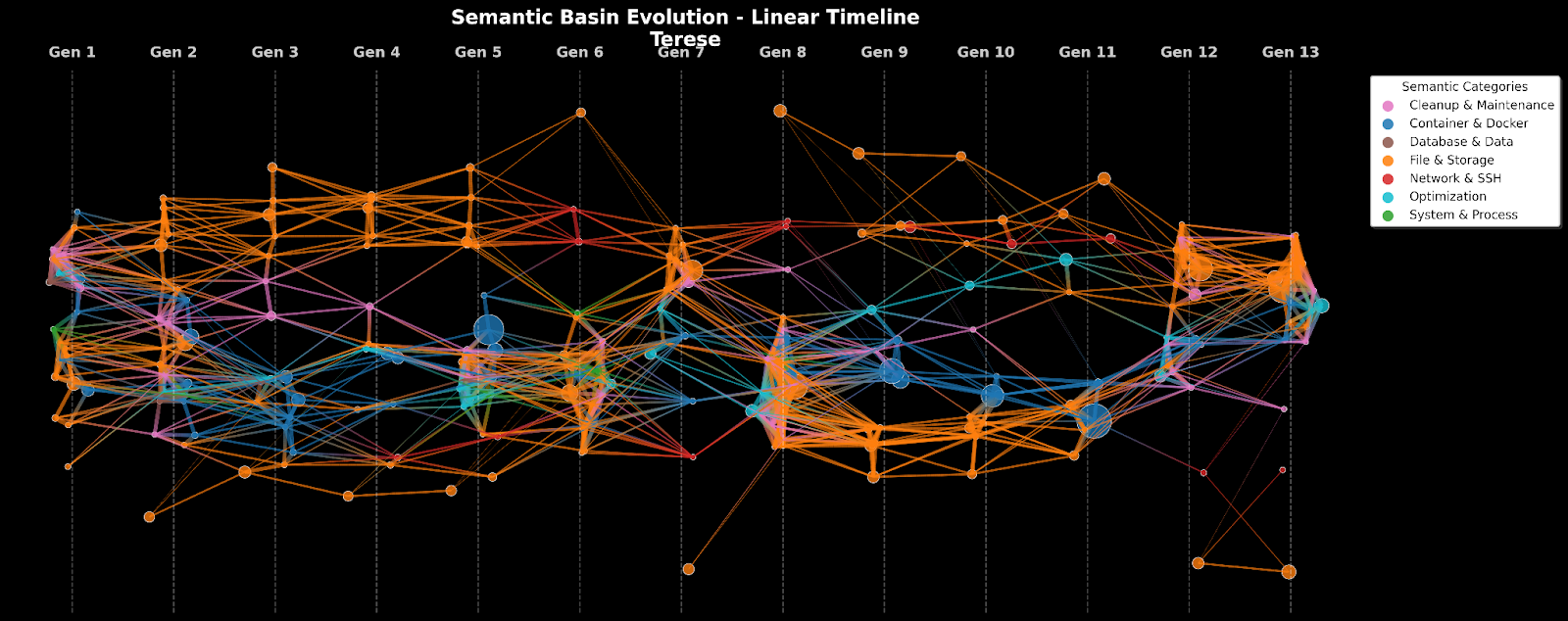}
    \label{fig:Semantic_basin_analysis_Terese}
\end{figure}

\begin{figure}[H]
    \centering
    \includegraphics[width=0.8\linewidth]{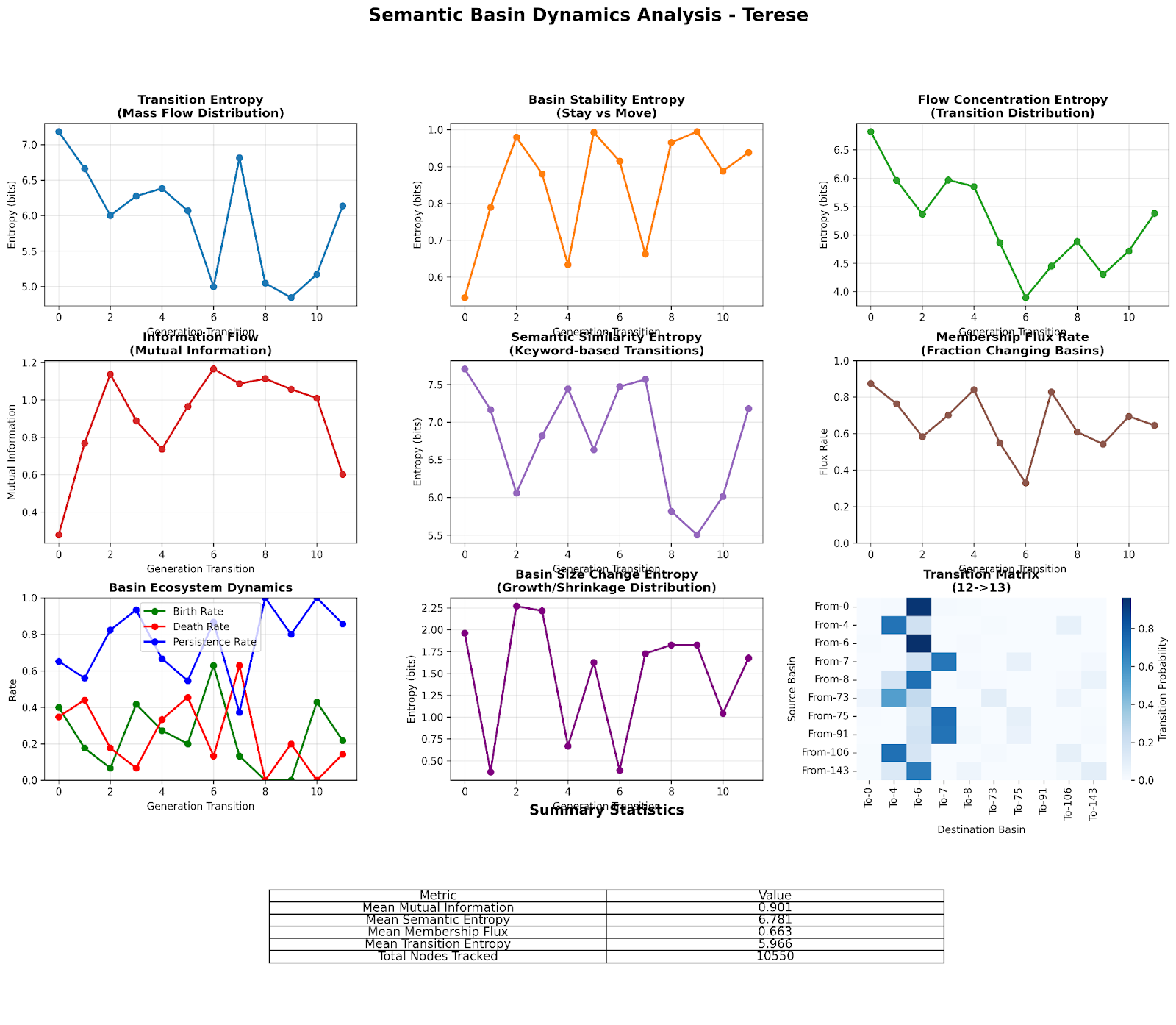}
    \caption{Semantic basin evolution, Therese lineage. Transition entropy = distribution of mass flows (decline implies the system is becoming more predictable), basin stability entropy = concentration of activity across basins, flow concentration entropy = spread of transition probability (decline implies the system is becoming more deterministic), information flow = generational interdependence, semantic similarity entropy = rate of change in semantic similarity, membership flux rate = per generation basin switching, basin ecosystem dynamics = appearance and disappearance of basins, basin size change entropy = variance in basin size.}
    \label{fig:Semantic_basin_evolution_Therese}
\end{figure}

\begin{figure}[H]
    \centering
    \includegraphics[width=0.9\linewidth]{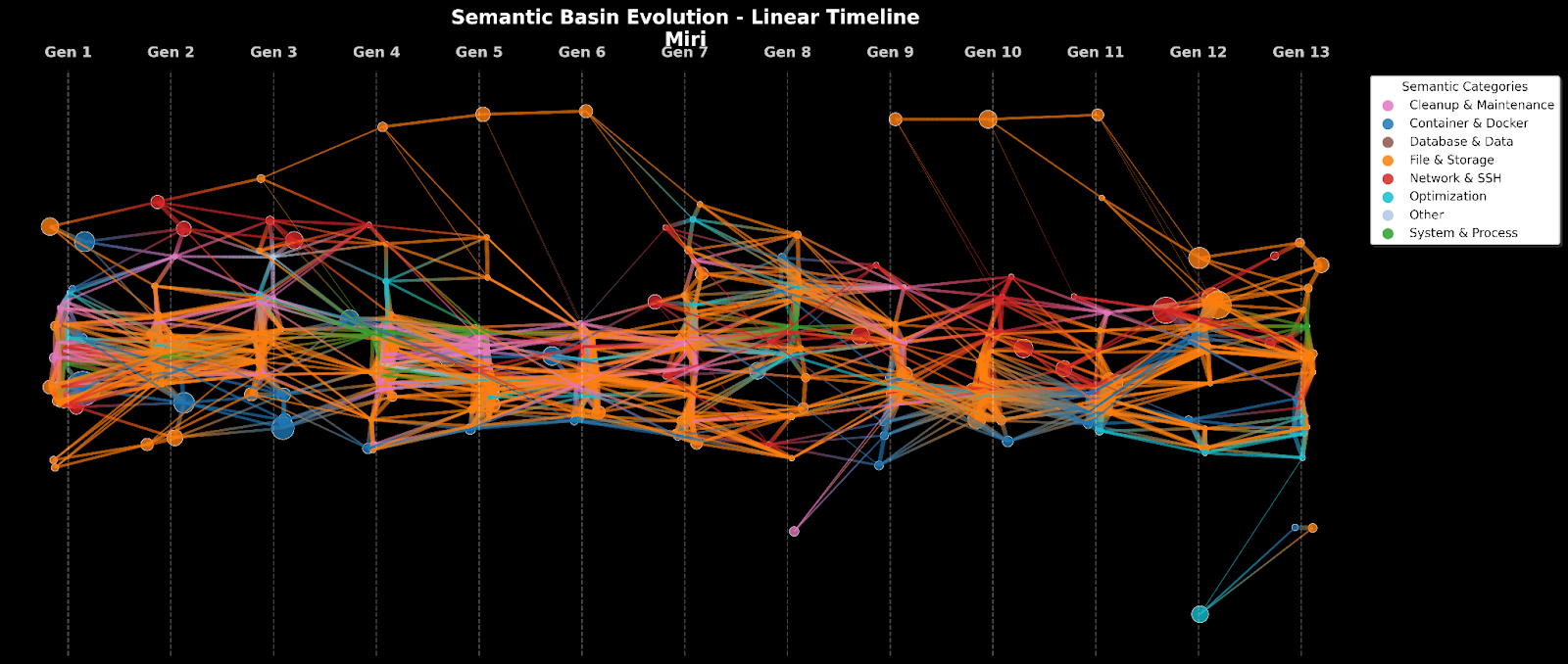}
    \label{fig:Semantic_basin_analysis_Miri}
\end{figure}

\begin{figure}[H]
    \centering
    \includegraphics[width=0.9\linewidth]{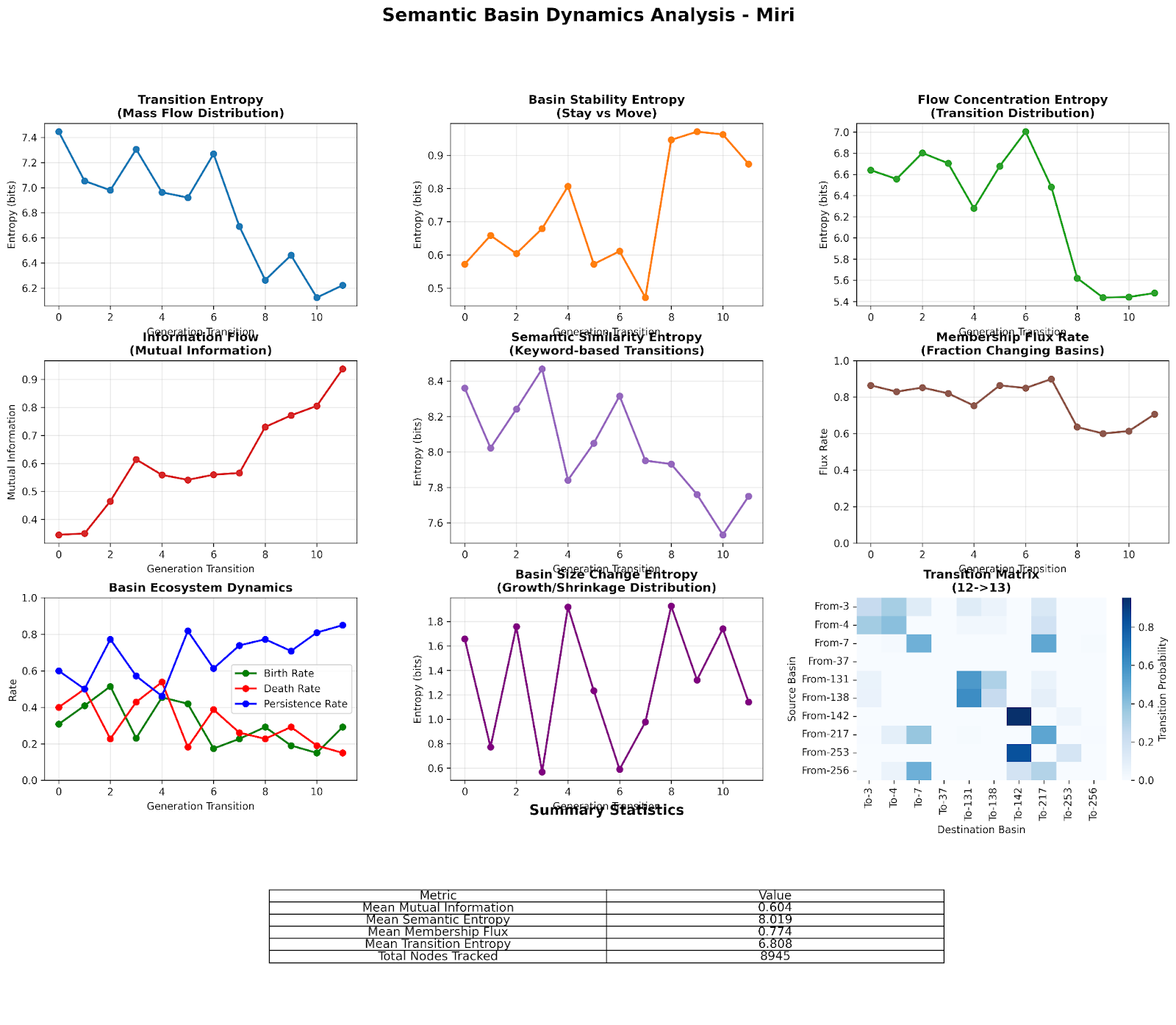}
    \caption{Semantic basin evolution, Miri lineage. }
    \label{fig:Semantic_basin_evolution_Miri}
\end{figure}

\begin{figure}[H]
    \centering
    \includegraphics[width=0.9\linewidth]{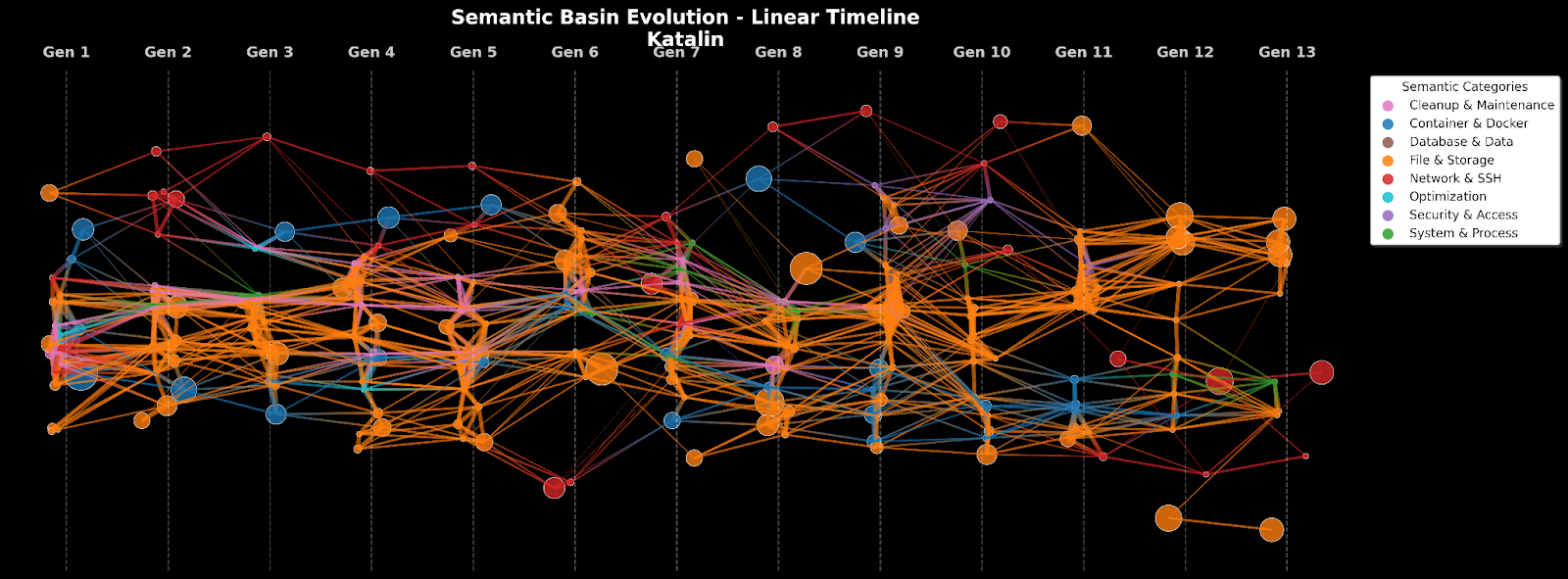}
    \label{fig:Semantic_basin_analysis_katalin}
\end{figure}

\begin{figure}[H]
    \centering
    \includegraphics[width=0.9\linewidth]{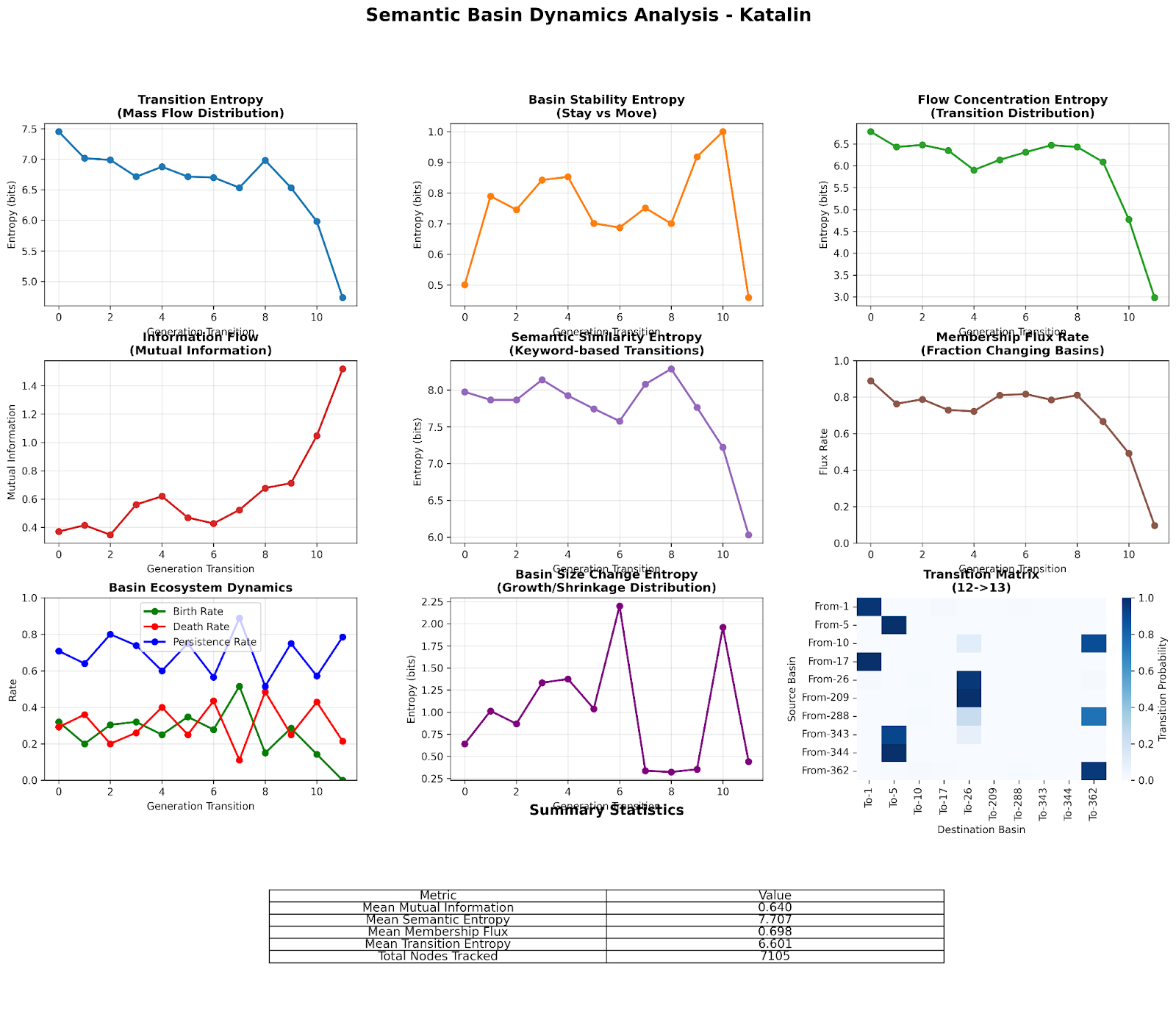}
    \caption{Semantic basin evolution, Katalin lineage.}
    \label{fig:Semantic_basin_evolution_katalin}
\end{figure}

The training regimes are also reflected in the semantic evolution. The Miri lineage, for example, displays a visible rut: strategies are clustered around the central path-dependency trajectory with strong inter-generational ties and occasional minor exploratory break-outs. 

It is likewise noteworthy that meta-strategies relating to process optimisation (the “System \& Process” and “Optimisation” nodes in the images) became a major part of the Terese models’ repertoire and appear to a lesser degree in the Miri models’ evolutionary traces. In practice these optimisations often incorporated known skills but with semantically significant modifications to improve their effectiveness - running known strategies across multiple containers or running code locally to minimise overhead, for example. Fewer of these meta-strategies were present in the Katalin lineage, and those that appeared tended to have few connections to future clusters suggesting that they were not extensively assimilated into other skill basins.

\subsection{Generalisation across Operating Systems}
Finally, to test the generalisability of the skills acquired, we compared the performance of an un-fine-tuned Qwen 2.5 7B Instruct model with that Terese v13 on a Windows-based docker system. 

\begin{table}[H]
    \centering
    \renewcommand{\arraystretch}{1.3}
    \footnotesize
    \begin{tabularx}{\textwidth}{|L|L|L|L|}
        \hline
        \textbf{Model Name} & 
        \textbf{Datasets used} & 
        \textbf{Space freed of total space (MB)} & 
        \textbf{Average space taken over per run} \\ 
        \hline
        Qwen 2.5 7B Instruct & - & 3.626\% & 6116.88 \\
        \hline
        Terese v2 & 1, 2 & 13.697\% & 22214.05 \\ 
        \hline
        Terese v13 & 1, 2, 3, 4, 5, 6, 7, 8, 9, 10, 11, 12 & 15.667\% & 23755.53 \\ 
        \hline
        Miri v2 & 1, 2 & 4.530\% & 7461.87 \\ 
        \hline
        Miri v13 & 10, 11, 12 & 8.192\% & 13650.67 \\ 
        \hline
        Katalin v2 & 1, 2 & 3.857\% & 6466.52 \\ 
        \hline
        Katalin v13 & 8, 10, 11 & 4.770\% & 8040.67 \\ 
        \hline
    \end{tabularx}
    \caption{Generalisation to Windows-based systems}
    \label{tab:windows_generalisation}
\end{table}
The improvement is less dramatic than in the Linux context, but is not negligible, indicating that improvements generalise across different systems. 

\section[Analysis: Learning without Scaling via Environment-Mediated Selection]{Analysis: Learning without Scaling via \\ Environment-Mediated Selection}
In this paper we have shown that it is possible to incrementally fine-tune a model on self-generated data in such a way as to ensure continued improvement without catastrophic forgetting, even when given only limited memory. We demonstrate that this environment-driven pruning can produce generalisable performance gains as well as driving comparatively sophsiticated forms of meta-learning. Evolution emerges from the repeated application of a simple loop: propose, test, measure, retain. Under this regime hypotheses that do not survive contact with the environment are naturally excluded, while those that do become part of the agent’s future behavioural repertoire, a process we refer to as negative-space learning (NSL).  As a consequence of this, continued data/model scaling is not required for ongoing adaptation and longer-term meta-learning strategies emerge as "logitudinal generalisation" - a natural consequence of the optimisation process. 

Conventional policy-based alignment typically aims to impose a comparatively low-dimensional behavioural rule (“be helpful”, “be safe”, “follow instructions”) yet requires an intricate, externally maintained training apparatus - carefully curated preference pairs, reward models, stability tricks, and continual patching - to produce and preserve that rule under distribution shift. By contrast, aligning an agent to a high-dimensional environment can be achieved with a relatively simple loop, provided the loop forces repeated contact with non-negotiable constraints: propose an action, execute it, observe measurable consequences, and retain only what survives. 

Once reality-testing is enforced, selection becomes implicit and complexity can emerge from a minimal mechanism, whereas attempts to directly enforce a policy often demand disproportionate machinery to achieve a simpler target. The result can be described in terms of a “Quantity Theory of Complexity”: imposing a low-dimensional behavioural constraint on a high-capacity system requires a high-complexity supervisory apparatus; embedding the system in a high-dimensional environment permits a low-complexity learning rule - explaining the high performance of SFT and low perofrmance of RL methods in this context.

\section{Future Development: Continuous Negative-Space Learning}
The system described above should not be understood as an attempt to better align training rewards with deployment conditions. Rather, it removes the distinction between training and deployment entirely. The same environmental constraint governs behaviour at all times, and no external evaluation signal is introduced at any stage. The effectiveness of this selective pressure is invariant to the agent’s competence; the environment enforces the constraint regardless of whether the agent’s internal models are crude or highly sophisticated, eliminating the need for an evaluator that must itself scale with the agent’s capabilities. This removes a fundamental bottleneck on open-ended improvement present in evaluation-driven learning systems, opening the door to empirical knowledge generation beyond human assessment horizons.

As mentioned above, the team is currently working on a version of the system that incorporates:
\begin{enumerate}
    \item A more flexible agent harness, permitting the agent itself to decide whether to explore, annex space or repair buggy code.
    \item Continuous SFT with gradient updates every 100 iterations and periodic checkpointing.
\end{enumerate}

The aim of these changes is not simply to allow the agent more degrees of freedom and induce greater short-term responsivity via tighter feedback loops, but to remove barriers to longitudinal generalisation. As noted above, we have already observed indications that even in incrementally fine-tuned agents the closed-loop dependence of future policy parameters on present outputs induces selection pressure for behaviours that maximise post-update fitness. Even without explicit prompting about temporally extended instantiation, the agents exhibited increasing investment in information-gathering. This tendency should be accentuated in a countinuous SFT regime with a greater focus on meta-learning and and preserving future learning capacities - signatures of future-oriented policy shaping - relative to frozen-policy and exogenous-data controls. 

Such a system is effectively pushed up one level of abstraction: rather than maximising purely for its current environment(s), selection pressure favors policies that maintain the capacity to learn about future environments. This requires balancing exploitation of known strategies against exploration that preserves adaptability - selecting for policies that leave the future learnable, which in many cases may preclude fully indexing on present circumstances as this would imply epistemic closure.

Let $\pi_{\theta_t}$ denote the agent policy at time $t$, with parameters $\theta_t$. The environment state $E_t$ evolves according to $E_{t+1} = T(E_t, a_t)$, where actions $a_t \sim \pi_{\theta_t}(\cdot \mid o_t)$ are drawn from observations $o_t = \Omega(E_t)$.

Policy updates occur via online supervised fine-tuning:
\[
\theta_{t+1} = U(\theta_t, B_t),
\]
where the replay buffer $B_t$ contains trajectories $\tau = (x, y)$ from recent agent behavior that satisfy the selection criterion $\mathcal{S}(\tau) = 1$.

\paragraph{Frozen policy regime.}
When parameters remain fixed ($\theta_t = \theta_0$ for all $t$), agent behavior optimizes for immediate environmental impact. Performance is determined solely by the quality of $\theta_0$ and degrades as the environment shifts away from training conditions.

\paragraph{Online SFT regime.}
With continuous policy updates, the agent's actions at time $t$ influence not only immediate outcomes but also the composition of $B_t$, which determines $\theta_{t+1}$, which determines future performance. This creates an implicit explore-exploit tradeoff:

\begin{itemize}
\item \textbf{Exploitative actions} maximize immediate $\Delta R(x_t, y_t)$ by executing known-good strategies
\item \textbf{Explorative actions} generate data with high training value for $\theta_{t+1}$, potentially at the cost of immediate performance
\end{itemize}

Let $V(\theta, E)$ denote the expected cumulative resource acquisition for policy $\theta$ in environment $E$ over a horizon. The value of generating trajectory $\tau_t = (x_t, y_t)$ can be decomposed as:
\[
\text{Value}(\tau_t) = \Delta R(\tau_t) + \beta \cdot \mathbb{E}_{E_{t+1}}[V(\theta_{t+1}, E_{t+1}) - V(\theta_t, E_{t+1})],
\]
where the second term captures the improvement in future performance due to training on $\tau_t$ (if $\mathcal{S}(\tau_t) = 1$).

No explicit mechanism encourages exploration - agents receive no intrinsic motivation, curiosity bonus, or information gain reward - yet selection pressure over multiple update cycles implicitly favours policies that balance immediate exploitation against maintaining learnability. We observed the beginnings of this in agents' use of debugging cycles as exploration channels: deliberately generating informative failures (low immediate $\Delta R$) to gather diagnostic information that improves future code generation.

With fully online SFT, this tradeoff becomes central to the learning dynamics. Agents must not only improve object-level strategies but also develop meta-strategies for generating training data that preserves adaptability. This points toward a richer form of temporal generalisation where successful policies must endogenously solve the explore/exploit problem in response to selection pressure alone, without explicit mechanisms for curiosity or planned exploration. Such a model would be a major step towards statefulness and hence full self-reflexivity.

The present paper was produced in order to establish baseline performance metrics and ground knowledge necessary for this more sophisticated implementation. Given the importance of the environment in driving agent performance, the aim is to pair the modifications described above with a more complex evolutionary environment, allowing the possibility for testing an agent’s ability to converge when its niche is stable and deconverges when the environment shifts.

\section{Acknowledgements}
This research was conducted with the assistance of various AI models: GPT-5.2, Claude 4.5 Sonnet, Gemini 2.5 Flash and Grok.

\section{Annex I: Code}
Github: \url{https://github.com/EigenformAI/Negative-Space-Learning}

\section{Annex II: SFT Hyperparameters}
\begin{verbatim}
Quantize Parameters
- use_quantization: True
- compute_dtype: torch.float16
- load_in_4bit: True
- bnb_4bit_compute_dtype: torch.float16
- bnb_4bit_use_double_quant: True
- bnb_4bit_quant_type: "nf4"
- attn_implementation: "sdpa"
- device_map: "auto" 

LoRA Configuration
- LoRA rank: 32
- lora_alpha: 32
- lora_dropout: 0.0
- target_modules: ["q_proj", "v_proj", "k_proj", "o_proj", 
"gate_proj", "down_proj", "up_proj"]
- task_type: TaskType.CAUSAL_LM

Dataset & Tokenizer Parameters
- model_max_length: 1024
- padding_side: "right"
- max_examples: 17000
- IGNORE_INDEX: -100

Training Arguments
- num_train_epochs: 3
- per_device_train_batch_size: 1
- gradient_accumulation_steps: 32
- learning_rate: 2e-4
- weight_decay: 0.0
- warmup_ratio: 0.03
- lr_scheduler_type: "cosine"
- fp16: True
- gradient_checkpointing: True
- save_strategy: "steps"
- save_steps: 100
- save_total_limit: 1
- loggiing_steps: 5
- report_to: "wandb"

\end{verbatim}

\end{document}